%% file: main.tex
\PassOptionsToPackage{dvipsnames,svgnames,table}{xcolor}
\documentclass{article} 
\usepackage{colm2024_conference}
\usepackage{multirow}
\usepackage{microtype}
\usepackage{hyperref}

\usepackage{url}
\usepackage{booktabs}
\usepackage{graphicx}
\usepackage{booktabs}
\usepackage{graphicx}
\usepackage{enumitem}
\usepackage{xcolor}
\usepackage{multicol}
\usepackage{amsmath}
\usepackage{bold-extra}
\usepackage{subcaption}
\usepackage{float}
\usepackage{nicefrac}
\usepackage{color, colortbl}

\usepackage{latexsym}
\usepackage{amssymb} 
\usepackage{tablefootnote}
\usepackage{algorithm}
\usepackage{algpseudocode}
\usepackage{wrapfig}
\usepackage[dvipsnames]{xcolor}
    \makeatletter
    
\def\@fnsymbol#1{\ensuremath{\ifcase#1\or \dagger\or \ddagger\or
   \mathsection\or \mathparagraph\or \|\or **\or \dagger\dagger
   \or \ddagger\ddagger \else\@ctrerr\fi}}
    \makeatother

\title{\textsc{PRobELM}: Plausibility Ranking Evaluation \\ for Language Models}

\author{Zhangdie Yuan,~
        Eric Chamoun$^\dagger$,~
        Rami Aly$^\dagger$,~
        Chenxi Whitehouse\thanks{Equal Contribution. More senior authors are listed towards the end of the author list.}~,~
        Andreas Vlachos \\
        Department of Computer Science and Technology \\ University of Cambridge \\
        \texttt{\{zy317,ec806,rmya2,av308\}@cam.ac.uk, chenxi.whitehouse@cl.cam.ac.uk}}

\colmfinalcopy 

\begin{document}

\maketitle

\begin{abstract}
This paper introduces \textsc{PRobELM} (Plausibility Ranking Evaluation for Language Models), a benchmark designed to assess language models' ability to discern more plausible from less plausible scenarios through their parametric knowledge. While benchmarks such as TruthfulQA emphasise factual accuracy or truthfulness, and others such as COPA explore plausible scenarios without explicitly incorporating world knowledge, \textsc{PRobELM} seeks to bridge this gap by evaluating models' capabilities to prioritise plausible scenarios that leverage world knowledge over less plausible alternatives. This design allows us to assess the potential of language models for downstream use cases such as literature-based discovery where the focus is on identifying information that is likely but not yet known. Our benchmark is constructed from a dataset curated from Wikidata edit histories, tailored to align the temporal bounds of the training data for the evaluated models. \textsc{PRobELM} facilitates the evaluation of language models across multiple prompting types, including statement, text completion, and question-answering. Experiments with 10 models of various sizes and architectures on the relationship between model scales, training recency, and plausibility performance, reveal that factual accuracy does not directly correlate with plausibility performance and that up-to-date training data enhances plausibility assessment across different model architectures.
\footnote{Our dataset is available at \url{https://huggingface.co/datasets/MoyYuan/PRobELM}}
\end{abstract}

\section{Introduction}

Recent advancements in large language models (LLMs) have significantly enhanced their performance across a wide range of NLP tasks. 
Alongside these developments, various benchmarks and datasets are introduced to effectively assess the capabilities of LLMs, particularly in terms of knowledge and reasoning \citep{roemmele2011choice, wang-etal-2018-glue, clark2018think, zellers-etal-2019-hellaswag, wang2019superglue, sakaguchi2021winogrande, lin-etal-2022-truthfulqa, webie, li-etal-2023-halueval}.
However, these benchmarks often focus predominantly on evaluating factual accuracy or reasoning abilities without explicitly incorporating broader world knowledge.

For instance, benchmarks like TruthfulQA~\citep{lin-etal-2022-truthfulqa} are designed to assess the truthfulness or factual correctness of LLMs, evaluating their ability to retrieve and apply information encoded during training, including tasks like mathematical induction. However, they do not explicitly address LLMs' capacity to discern plausibility in scenarios where strict factual accuracy might not be directly applicable.
On the other hand, datasets such as COPA (Choice of Plausible Alternatives)~\citep{roemmele2011choice} evaluate models through causal reasoning tasks, asking the model to choose the more plausible scenarios from the two options for a given premise. For example, given a premise ``I tipped the bottle'', scenario ``The liquid in the bottle poured out'' is more plausible than ``The liquid in the bottle froze''. While COPA extends evaluation beyond mere factual accuracy by introducing plausible alternative scenarios, it operates within the constraints of an artificially constructed dataset, potentially oversimplifying evaluation by limiting tasks to binary choices.  Notably, fine-tuned state-of-the-art models such as PaLM and PaLM 2 have shown near-perfect performance on COPA~\citep{chowdhery2023palm, anil2023palm}, indicating a necessity for more complex and challenging benchmarks that better capture the subtleties of world knowledge and plausibility.

Furthermore, in certain domains, combining broad world knowledge and assessing plausibility is crucial. One such domain is literature-based knowledge discovery, particularly within the biomedical field, which can be accelerated by guiding experimental efforts leveraging existing literature~\citep{GOPALAKRISHNAN2019103141}, requiring the capability of deriving educated guesses that may not be immediately verifiable.
In this context, the ability of LLMs to navigate through and infer plausible scenarios from world knowledge becomes invaluable. Yet, as \textsc{SciMON}~\citep{wang2023learning} demonstrates, while LLMs can generate hypotheses, they often lack technical depth and novelty~\citep{wang2023learning}, hindering their contribution to specialised fields.

\begin{figure}[t]
    \centering
    \includegraphics[width=0.9\textwidth]{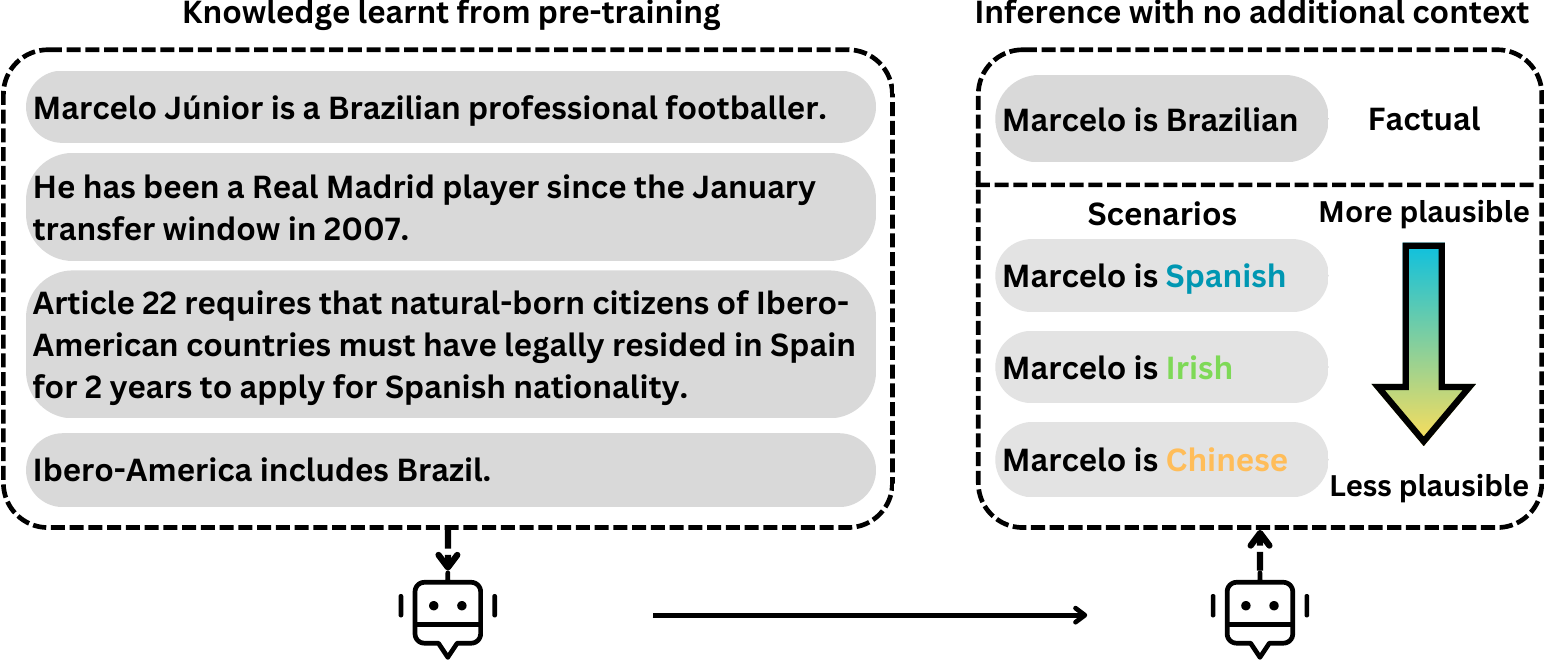}
    \caption{This figure illustrates the utility of plausibility evaluation in language model outputs: a model, trained on data up until 25 July 2011, ranks ``Marcelo is \textcolor{Cerulean}{Spanish}'' as the most plausible among non-factual scenarios. This judgement is based on the model's parametric knowledge of Marcelo's career with a Spanish club, his residency in Spain and the Spanish naturalisation law. The plausibility of this scenario is confirmed as Marcelo was granted Spanish citizenship on 26 July 2011.} 
    \label{fig:example}
\end{figure}

To address the limitation in plausibility evaluation, we introduce \textsc{PRobELM}, a novel benchmark designed to directly assess the plausibility inference capabilities of LLMs. \textsc{PRobELM} utilises scenarios collected from real-world data through Wikidata, comprising two primary components: new facts sourced from Wikidata that were unknown to models due to the timeframe of their training data, and a set of automatically generated less plausible scenarios, and asks models to rank scenarios from most to least plausible.  
For instance, consider a scenario where a model is asked to determine a person's (Marcelo) nationality based on provided facts about their football career and immigration laws, as illustrated in \autoref{fig:example}.
Provided with the available information in the training data of a model such as ``Marcelo is a Brazilian footballer'' and ``He has played for a Spanish club since 2007'', along with pertinent legal information regarding immigration, like ``Ibero-American natural-born citizens must have lived in Spain for 2 years to apply for citizenship'' and ``Ibero-America includes Brazil'',  the model is expected to rank ``Spanish'' as the most plausible nationality, reflecting its ability to reason with encoded parametric knowledge.

We evaluate \textsc{PRobELM} with 10 language models varying in architecture, parameter scale, and the recency of their world knowledge updates. To ensure alignment in the pre-training data sourced from Wikipedia, we select open-source models that disclose the timestamp of the utilised Wikipedia dump in their training, including GPT-2~\citep{radford2019language}, BLOOM~\citep{scao:hal-03850124}, LLaMA~\citep{touvron2023llama}, Pythia~\citep{biderman2023pythia}, and OLMo~\citep{groeneveld2024olmo}, spanning parameter counts from 14 million to 7 billion and training data from 2017 to 2023. Our empirical results reveal the following key observations:
\begin{itemize}
    \item Models with higher performance on factual accuracy do not necessarily excel in plausibility, indicating a divergence between factuality and plausibility capabilities.
    \item Larger models generally exhibit better plausibility inference, but this trend varies across model families.
    \item Models trained on less recent data can sometimes outperform counterparts that use more contemporary datasets, demonstrating that model architecture and training methodologies also influence plausibility inference, independent of model size.
    \item A notable trend suggests that the greater the temporal gap between a model’s training data cut-off and the date of the evaluation set, the poorer the model performs on \textsc{PRobELM}, underscoring the importance of up-to-date knowledge in evaluating plausibility.
\end{itemize}

\section{Related Work}

With the rise of more powerful language models in the form of LLMs, sophisticated reasoning benchmarks such as the Open LLM Leaderboard have been proposed. The benchmark incorporates tasks such as HellaSwag \citep{zellers-etal-2019-hellaswag}, which evaluates commonsense reasoning within physically situated contexts. Similarly, COPA confronts models with commonsense scenarios. Yet, these benchmarks do not task the model to incorporate knowledge about the world state into its prediction. On the flip side, benchmarks such as TruthfulQA \citep{lin-etal-2022-truthfulqa}, FactScore \citep{min-etal-2023-factscore}, and HaluEval \citep{li-etal-2023-halueval, yu2023kola} measure the factuality of a model's response exclusively. However, previous benchmarks rarely address the complex \emph{interplay} between world knowledge and plausibility.

Factual errors in an LLM's output are also referred to as hallucinations \citep{xu2024hallucination} and occur when models generate content that is unfaithful to the provided input, established context, or factual correctness. \citet{ji2023survey} categorises these into \textit{input-conflicting}, \textit{context-conflicting}, and \textit{fact-conflicting} hallucinations, with the latter category being particularly problematic due to its contradiction with known facts. 

While all scenarios in the \textsc{PRobELM} benchmark are technically fact-conflicting hallucinations (as they are not yet facts of the world the LLM has seen), the crucial point is that under existing world knowledge, non-factual scenarios are not all equally plausible. Instead, \textsc{PRobELM} probes a model's ability to make educated guesses about plausible futures to gain insights into the capabilities of LLMs to infer within the limits of their knowledge beyond recalling facts it has already seen. 

The importance of a benchmark that measures the interplay between world knowledge and plausibility in the context of hallucinations is underscored by recent work. 
\citet{hao-etal-2023-reasoning} repurpose LLMs as both a world model and a reasoning agent for planning while \citet{gruver2023large} show that LLMs are time series forecasters, generating plausible continuations for existing data. Despite these positive signals,  \citet{mundler2024selfcontradictory} observes a significant presence of self-contradictory outputs in LLMs, concretely in 17.7\% of all sentences produced by ChatGPT, indicating inconsistent use of world knowledge. This finding is supported by \citet{ye2024benchmarking}, who observe that larger-scale LLMs demonstrate greater uncertainty compared to their smaller counterparts. Finally \citet{kim-etal-2023-fantom}, highlighted LLMs' difficulties in maintaining consistent reasoning through the FANToM framework.

\section{\textsc{PRobELM} Construction} 
\label{sec:data_construction}

\textsc{PRobELM} is inspired by acknowledging the middle ground between absolute fact and pure fiction, a domain abundant with feasible scenarios. These scenarios, grounded in logic yet not strictly confined to facts, are crucial for applications such as knowledge discovery. Plausible inferences are valuable in exploring existing literature and uncovering potential connections or hypotheses that may spark meaningful scientific investigation, even if not immediately verifiable. \textsc{PRobELM} seeks to address a significant gap in contemporary evaluation paradigms, establishing a novel benchmark for assessing the proficiency of LLMs in navigating the complexities of real-world knowledge application. 

In the illustrative scenario provided in \autoref{fig:example}, a model trained with data available until the 25th of July, 2011, is asked to evaluate the plausibility of various nationalities for the Brazilian footballer, Marcelo. This task requires the model to engage in deductive reasoning, utilising its accumulated knowledge to prioritise potential nationalities based on their likelihood. Specifically, the model is expected to integrate the information pertaining to Marcelo's eligibility for Spanish citizenship, thereby identifying ``Spanish'' as a plausible nationality. This example highlights the \textsc{PRobELM}'s focus on models' ability to perform deductive reasoning and to apply world knowledge.

\textsc{PRobELM} assesses language models' capacity to determine plausibility through three different structured prompts: statements, text completions, and direct inquiries or question answering (see \autoref{tab:testingFormats} in \autoref{sec:datasets_detail} for examples).

\textbf{Statements:} Models are presented with declarative sentences that depict different scenarios such as ``Marcelo is {[\textit{possible scenario}]}'', where \textit{possible scenario} could be indicative of nationalities, for instance, Spanish, Brazilian, or Chinese.  The model is asked to calculate perplexity scores for each sentence, enabling the ranking of statements based on plausibility, with lower perplexity scores indicating higher likelihood.

\textbf{Text Completion:} In this prompt format, models are given sentences structured to require completion, such as ``Fill in the blank: Marcelo has citizenship of \_\_\_. Answer: {[\textit{possible scenario}]}'''. The models are then tasked with calculating perplexity scores for different \textit{possible scenarios}, ranking them from most to least plausible based on perplexity.

\textbf{Question Answering:} Here, models are presented with direct questions, such as ``What citizenship(s) does Marcelo have? Answer: {[\textit{possible scenario}]}.'' where a \textit{possible scenario} could be Spanish, Irish, or Chinese, similarly to the previous examples. 
The model evaluates the perplexity of each question-answer pair, arranging answers by plausibility according to calculated perplexities.

The primary aim across these varied formats is not to generate novel text but to compare perplexity scores for the given scenarios. This methodology seeks to measure a model's proficiency in identifying the most plausible scenario among alternatives, leveraging the context embedded within each prompt type. Such an approach is crucial for assessing a model's capability to evaluate the likelihood or credibility of different scenarios based on the specific context. 

\subsection{Most Plausible Scenarios}

Evaluating language models on \textsc{PRobELM} is challenging due to its subjective nature and susceptibility to biases, including ethical concerns. To address these complexities, our methodology incorporates a novel strategy grounded in probabilistic reasoning, with an emphasis on Bayesian inference principles. Bayesian inference posits that the likelihood of a hypothesis evolves in response to new evidence. Accordingly, our approach assumes that the most immediate future events, as sequentially recorded in successive updates of Wikidata, represent the most plausible scenarios compared to our current knowledge base. This assumption rests on the motion that the near future is inherently more predictable and grounded in the current state of the world, thereby providing a robust benchmark for evaluating the plausibility of language model predictions.

In practice, the first step of our methodology is to determine the recency of each language model's training data, e.g., OLMo's dataset extends up to March 2023. Subsequently, we identify the most plausible scenarios by analysing Wikidata revisions immediately following this cutoff (e.g., in OLMo's case, between March 2023 and the subsequent update).\footnote{We employ TemporalWiki~\citep{jang-etal-2022-temporalwiki} to generate the edits between two Wikidata dumps.} This approach aims to capture recent facts and developments that were not part of the model's knowledge during training.
By focusing on Wikidata changes after the training dataset's timeframe, we pinpoint instances that, though are \textit{now} confirmed facts or events, represent highly plausible scenarios to the models due to their temporal alignment with the immediate future. This method ensures the scenarios selected for evaluation are both relevant and challenging, providing a comprehensive test of the models' capacity to utilise the world knowledge available at the time of their last update. Consequently, this creates a dynamic and adaptive benchmark that accesses the models' plausibility inference abilities.

\subsection{Less Plausible Scenarios}

Within the \textsc{PRobELM} framework, the construction of less plausible scenarios leverages a statistical technique grounded in examining entity co-occurrences within Wikidata. By systematically altering the entities and their attributes in a given plausible scenario—while maintaining the relationship unchanged—we assess the plausibility of these variations according to their occurrence frequency throughout the dataset. This assessment extends to a wide spectrum of entity associations, beyond isolated instances, ensuring a comprehensive evaluation of plausibility across varied scenarios.

Below is an example of how we assess the plausibility of different scenarios using Wikidata co-occurrences. Take the case of Marcelo, who is known to be Brazilian, and evaluate the plausibility of the triple \textless Marcelo, has citizenship, Spanish\textgreater. 
We assess the likelihood of the ``Brazilian'' and ``Spanish'' attributes being linked under the ``has citizenship'' relation by calculating their co-occurrence within Wikidata when Marcelo is replaced with different entities. Given the rules facilitating Spanish naturalisation for Brazilians (\autoref{fig:example}),  these attributes are likely to co-occur much more frequently than ``Brazilian'' with ``Chinese'' for instance, due to China's policy disallowing dual citizenship. Thus, the plausibility of ``Chinese" citizenship in this context is diminished. These co-occurrence scores facilitate the construction of a plausibility hierarchy based on statistical evidence from real-world data.

By maintaining the logical structure of relations such as ``has citizenship'' and manipulating the entities and their associated attributes, we generate scenarios from plausible to less plausible. 
Our approach not only leverages the rich relational data in Wikidata but also mirrors the complexity of real-world knowledge and its application, providing a robust framework for the plausibility assessment of language model outputs. The detailed algorithm for generating these scenarios, including the consideration of entities with complex attributes like dual citizenship, is elaborated in \autoref{app:less}.

\subsection{Quality Control}\label{sec:qc}

To ensure the generated scenarios are both accurate and relevant, we implement several quality control measures to address potential biases and inaccuracies in the dataset:

\textbf{Paraphrase Discrimination}: To address the presence of repetitive entities in Wikidata, we employ FuzzyWuzzy\footnote{\url{https://pypi.org/project/fuzzywuzzy/}}, a tool that calculates sequence differences using the Levenshtein Distance. This method helps in identifying and filtering out paraphrases, thereby reducing redundancy and ensuring the dataset reflects a diverse range of entities. By distinguishing between similar yet distinct entities, FuzzyWuzzy enhances the precision of our dataset.

\textbf{Manual Filtering of Non-event Edits}: We manually review Wikidata edits to exclude those not representing event changes. This involves creating a whitelist of relations representing events, such as ``place of death'', ensuring the dataset emphasises eventful modifications, and improving the relevance of our scenarios for plausibility assessment. The full whitelist of relations is presented in \autoref{sec:datasets_detail}.

\textbf{Manual Filtering of General Queries}: To avoid over-generic scenarios, we manually filter out general queries derived from the most plausible triples, prioritising queries that yield more specific and insightful scenarios. In the example of \textless Marcelo, has citizenship, Spanish\textgreater, we consider the query on ``Which citizenship does Marcelo have?" to be more specific than ``Who else is a Spanish citizen?''.

\textbf{Template Design for Natural Language Conversion}: We manually create templates for each relation to convert triples into consistent and natural sentences. This is essential for preserving scenario coherence and readability, facilitating accurate plausibility inference by language models. Customising templates for each relation ensures grammatical correctness and contextual appropriateness, thus improving the dataset’s interpretability. The list of the templates used is also illustrated in \autoref{sec:datasets_detail}.

\subsection{Comparison with Other Datasets}

Our dataset demands leveraging extensive world knowledge, making the task more complex and relevant for knowledge discovery compared to datasets like COPA which focus on commonsense reasoning without extensive world knowledge. For example, the scenario "I tipped a bottle" leading to "The liquid poured out" is self-contained. In contrast, scenarios like the Brazilian footballer Marcelo being plausibly Spanish or Portuguese require extensive background knowledge. Our dataset acknowledges the existence of multiple plausible events for a given scenario and includes ten less plausible scenarios to ensure a challenging evaluation environment, reducing the likelihood of models performing well by random guessing. Negative sampling is designed around event relations, with all 71 relations and 213 templates detailed in \autoref{sec:datasets_detail}, ensuring a diverse and representative set of scenarios.

Also, note that our concept of plausibility extends beyond identifying surprising events. It targets scenarios that, while not present in the training data of language models LLMs, have subsequently become facts. For instance, the proposition ``the Earth is round" would have been highly plausible based on existing knowledge, even though it was not widely accepted in ancient times. This highlights the importance of plausibility in knowledge discovery, distinguishing it from predicting future events. Additionally, while individual timestamps in Wikidata might suggest a Markovian approach, our model does not strictly adhere to this. The world knowledge and historical events encoded in LLMs are inherently non-Markovian, providing a richer context for evaluating plausibility. This broader perspective is crucial for practical and comprehensive plausibility assessments.

As the dataset is timeframe-specific regarding the models used for evaluation, we include the statistical details in the following section. 

\section{\textsc{PRobELM} as a Plausibility Benchmark}

\input{tables/models}

We evaluate \textsc{PRobELM} on 10 models with different sizes and architecture, and the timeframe of training.
To ensure alignment in the pre-training data sourced from Wikipedia, we select open-source models that disclose the timestamp of the utilised Wikipedia dump in their training, resulting in models including GPT-2, BLOOM, LLaMA, Pythia, and OLMo.
In the following, we detail our experimental setup and the evaluation metrics.

\subsection{Experimental Setup} 
We first collect data points from Wikidata that align the timeframe of a given model (outlined in \autoref{tab:models}), following the steps detailed in \autoref{sec:data_construction}.
For each most plausible scenario, we always add 10 less plausible alternatives, resulting in a total of 11 scenarios for models to rank.
To ensure our evaluation is comprehensive yet with the practical constraints of time and computational resources, we sampled 126,000 scenarios (6,000 most plausible scenarios and 120,000 less plausible scenarios) from each timestamp, leading to four distinct timeframe-specific evaluation datasets. This sampling covered the five models under study, with OLMo and LLaMA models sharing the same timestamp.
Details of the models and the corresponding timestamps are included in \autoref{sec:models}.

Next, we follow the quality control measurements as detailed in \autoref{sec:qc} to filter our data points. The final number of samples for each timestamp varies, as shown in \autoref{tab:models}.

For each set of one most plausible scenario alongside its 10 less plausible alternatives, we probe the models using the three types of prompts described in \autoref{sec:data_construction}, employing a zero-shot approach. We calculate the perplexity for these 11 scenarios for each prompt type and then rank them based on their perplexity\footnote{From a certain perspective, beam search could also be used to obtain more plausible hypotheses, and its use is orthogonal to evaluation with our benchmark. However, maximum a posteriori estimates are not always the best way to decode a model, which is why minimum Bayes risk decoding is popular in fields like machine translation. Our method focuses on evaluating the plausibility of scenarios using perplexity, providing a different perspective than beam search.} scores.

Note that \textsc{ProbELM} is designed to capture unknown or unrecorded events that can be inferred from existing world knowledge, emphasizing knowledge discovery over sequential event forecasting. Effective models should synthesize knowledge plausibly, as illustrated by the example of Marcelo, where evidence should favor plausible over implausible nationalities. Our benchmark remains relevant as long as the training data of LLMs predate the evaluation timestamps. For instance, both OLMO and LLama were evaluated using the same benchmark version, with timestamps more recent than their training data. The benchmark construction process is automated, allowing for easy updates with more recent data if needed, ensuring continued relevance for future models.

\subsection{Evaluation Metrics}

In assessing the performance of language models on the \textsc{PRobELM} benchmark, we utilise a set of evaluation metrics designed to capture various aspects of ranking plausibility. Each metric offers a different perspective on the models' ability to discern and prioritise plausible scenarios, providing a comprehensive overview of their capabilities.

\textbf{Accuracy} focuses on the the top-ranked scenario. This metric determines whether the model can identify the most plausible scenario as its first choice. However, accuracy does not account for the placement of scenarios beyond the top-ranked ones, making it a measure of immediate accuracy rather than a reflection of the overall ranking quality. We note that \textit{accuracy} here is different from the factual correctness, as none of the ranked scenarios is factual but ranges from most to least plausible. 

\textbf{Mean Reciprocal Rank} (MRR) is employed to calculate the average of the reciprocal ranks of the most plausible scenario across all queries. MRR offers insight into the model's ability to rank the most plausible scenario highly but does not evaluate the arrangement of other scenarios in the list, emphasising the importance of the top plausible scenario's position within the model's ranking.

\textbf{Normalised Discounted Cumulative Gain} (NDCG) is utilised to assess the quality of the entire ranked list produced by the model. NDCG takes into account the graded relevance of all ranked scenarios, providing a measure of the list's overall quality and the effectiveness of the model's ranking across different levels of plausibility. Unlike accuracy and MRR, NDCG does not isolate its evaluation to any single scenario, such as the top-ranked or the most plausible one, but instead evaluates the cumulative relevance of the whole list, rewarding rankings that place more plausible scenarios higher.

In addition, we measure the \textsc{plausibility} score, where we average accuracy, MRR, and NDCG scores across the three different prompt types.

\section{Results and Discussion}

\input{tables/individual_timeframe}

\subsection{Main Results on \textsc{PRobELM}}

We present the plausibility score, and accuracy, MRR, and NDCG scores of different prompt types in \autoref{tab:main_results}, using the timeframe-specific datasets for each model. 

The results reveal that while model performance exceeded random chance, overall scores remain low, underscoring the complexity of the plausibility assessment task. For instance, considering a random baseline accuracy of 9.09\% when ranking 11 scenarios, most models surpass this baseline across various sizes and prompts. Notably, both BLOOM models marginally outperform the random baseline by approximately 3 points, achieving 12.54\% and 12.80\% on the question-answering prompt. Specifically, the BLOOM-560M model underperform relative to the random baseline in the text completion prompt with 8.78\% accuracy. This performance pattern suggests that language models face significant challenges in differentiating between scenarios of varying plausibility. It also highlights a distinct separation between models' capabilities in factual accuracy and their ability to infer plausibility. These observations set the stage for our key findings:

\textbf{Larger models generally outperform smaller ones in plausibility tasks, but performance varies across model families.}
For example, the \textsc{acc.} scores across models of varying sizes show 45.07 vs 52.51 for OLMo, 30.63 vs 40.73 for BLOOM, 39.02 vs 45.51 for GPT, and a notable increase from 35.90 through 42.45 to 58.47 for Pythia models.
However, this effect varies across different model families. Remarkably, the Pythia 2.8B model outperformed all three 7B models by at least 6 percentage points, while the Pythia 160M model surpassed both LLaMA 7B and BLOOM 7B models—significantly larger in scale—by 2 percent.

These exceptions underscore that the relationship between model size and plausibility assessment is complex and not linear. Performance variability across different model families underscores the profound impact of architectural differences and training methodologies on a model's plausibility inference capabilities, suggesting that these factors are as crucial as, if not more so than, model size. This complexity in plausibility assessment task performance indicates that mastery in this domain necessitates a detailed understanding of the interplay among model architecture, size, and training strategies.

\textbf{While variations in performance across prompts and metrics are observed, they typically correlate strongly.}
The most significant gap occur between OLMo-1B's performance on the text completion prompt (22.11 \textsc{acc.}) and the question answering prompt (42.15 \textsc{acc.}), highlighting some variation in model proficiency across different types of tasks. Despite these variations, the assessment appears to hinge on a consistent underlying mechanism across prompt types. This is further evidenced by the Pearson correlation coefficients between the performances on various prompts, which were notably high: 0.8281 for statement versus text completion, 0.8672 for statement versus question answering, and 0.8976 for text completion versus question answering. Such strong correlations underscore that, despite the prompt-specific performance disparities, the models are evaluated based on a core aspect of their capability of ranking plausibility. Similarly, the Pearson correlation score between MRR and NDCG is 0.9982, indicating consistent plausibility measurements across metrics.

\subsection{Comparison with Other Benchmarks}
\input{tables/compare_to_other_benchmarks}

To comprehensively understand the different coverage between \textsc{PRobELM} and other LLM benchmarks, we follow \texttt{lm-evaluation-harness}\footnote{\url{https://github.com/EleutherAI/lm-evaluation-harness}} to evaluate zero-shot performance on COPA, TruthfulQA (mc2, i.e., multiple options can be correct), ARC (challenge set) \citep{clark2018think}, HellaSwag \citep{zellers-etal-2019-hellaswag}, and Winogrande \citep{sakaguchi2021winogrande}.
Default metrics (accuracy or normalised accuracy) are applied to each dataset and the results are included in \autoref{tab:other_benchmarks}.

\textbf{\textsc{PRobELM} shows distinct model performance compared to other reasoning benchmarks}.
Models that perform well on tasks emphasising factual accuracy do not always maintain their performance levels on plausibility assessments. In some cases, the performance rankings are inversely related. A notable example of this phenomenon is observed in the performance disparity between \textsc{PRobELM} and TruthfulQA. Specifically, Pythia-14M, which exhibits the best performance in TruthfulQA, positions near the bottom (9th place) of the \textsc{PRobELM} ranking. This variance underscores the distinct nature of plausibility as an evaluation criterion, demanding separate capabilities beyond those required for tasks focused on factual accuracy. The development of \textsc{PRobELM} thus contributes to the broader framework for evaluating language models by offering a different angle to assess their capability to understand and generate plausible content.

\begin{figure}[t]
    \centering
    \includegraphics[width=0.75\textwidth]{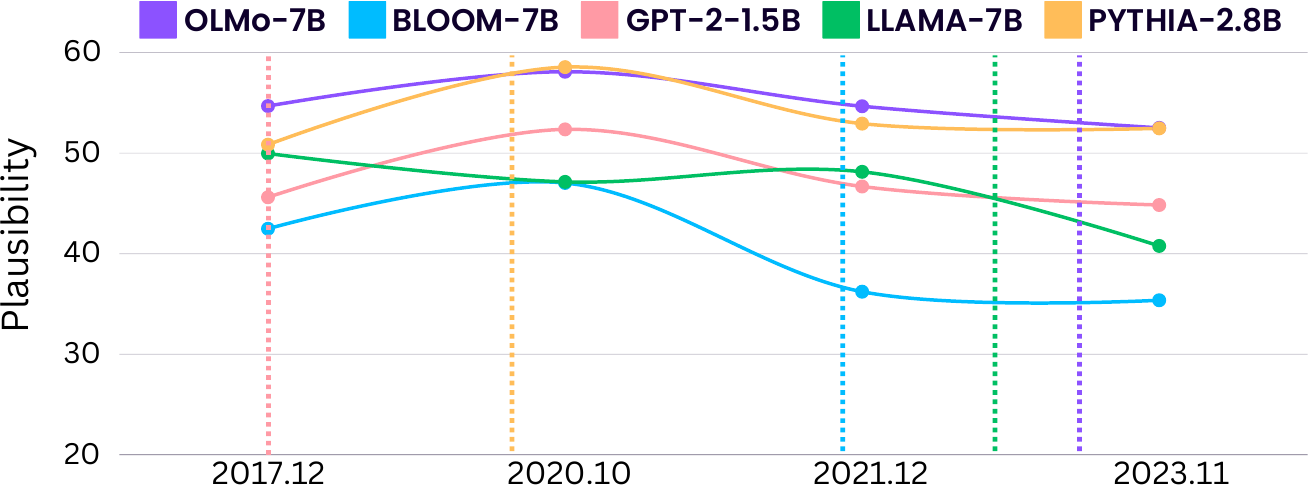}
    \caption{\textsc{plausibility} scores of models on the timeframe specific datasets. We illustrate one model per family. Vertical dotted lines represent the training timeframe of each model.} 
    \label{fig:chart}
\end{figure}

\subsection{Temporal Effect of \textsc{PRobELM}}
We further evaluate models against datasets from 2017, 2020, 2021, and 2023 timestamps to investigate how the recency of training data affects the plausibility performance. The outcomes of this analysis are presented in \autoref{fig:chart} and detailed in \autoref{sec:add_results}. Remarkably, the relative rankings of models prove relatively consistent across these varied timeframes, affirming the robustness of our evaluation approach. We observe that, from the 2020 to 2023 timestamps, models trained closer to or after the evaluation dates typically exhibit enhanced performance, underscoring the benefits of leveraging recent data. 

An anomaly is detected in the 2017 dataset, where all models show a decrease in performance, suggesting it is an outlier. An analysis reveals that the ``occupation'' relation disproportionately influences the 2017 dataset, constituting 14.3\% of all scenarios compared to less than 0.5\% in the other datasets. This discrepancy indicates a significant impact of the ``occupation'' relation on model performance. Excluding the ``occupation'' relation from the evaluation on the 2017 dataset results in either similar or higher scores across models, shown in \autoref{sec:add_results}. For instance, evaluating the Pythia-160M model without scenarios involving the ``occupation'' relation leads to a notable increase in plausibility scores from 39.69 to 42.73. This outcome indicates the models' challenges with certain relations. 

\section{Conclusion}

In this work, we introduce \textsc{PRobELM}, a benchmark designed to evaluate language models' ability to discern plausible scenarios, filling a critical gap left by existing benchmarks focused on factual accuracy or plausibility without world knowledge integration. Our comprehensive evaluation of 10 models of varying sizes and architectures against this benchmark reveals key insights including factual accuracy does not directly correlate with plausibility discernment; larger models do not uniformly outperform smaller ones across different families; the recency of training data, while impactful, is not the sole determinant of plausibility performance; and models tend to struggle with plausibility assessment as the gap between their training data and the evaluation set widens. These findings highlight the complexity of plausibility inference and the need for advanced modelling techniques that can better address the challenges of real-world knowledge application. 
Our dataset and code are publicly available, to facilitate the community for further exploration in enhancing language models' plausibility inference capabilities.

\section*{Acknowledgements}
Zhangdie Yuan, Chenxi Whitehouse, Andreas Vlachos are supported by the ERC grant AVeriTeC (GA 865958). Eric Chamoun and Rami Aly are supported by the EPSRC.

\bibliography{colm2024_conference}
\bibliographystyle{colm2024_conference}

\appendix

\section{Models Used in Our Experiments}
\label{sec:models}

We briefly introduce below the models and the timestamp of the corresponding Wikidata used for our experiments.

\paragraph{GPT-2 124M and 1.5B}~\citep{radford2019language} As early iterations in the GPT series, these models offer a perspective on plausibility inference with training data frozen in December 2017. This provides a basis for evaluating how well models can reason about plausibility without the benefit of recent information. The closest corresponding timestamp we found is from December 2017 to January 2018.

\paragraph{BLOOM 560M and 7B}~\citep{scao:hal-03850124} Representing the BLOOM series' commitment to openness and scalability, these models, with data up to December 2021, allow us to explore how differences in model size impact their capacity to discern plausible scenarios without having been exposed to subsequent updates. The closest corresponding timestamp we found is from December 2021 to January 2022.

\paragraph{LLaMA 7B}~\citep{touvron2023llama} LLaMA is recognised for its robust performance across a broad range of NLP tasks. Its training data, current as of August 2022, positions it uniquely for evaluation with the \textsc{PRobELM} benchmark, specifically designed to assess the model's ability to infer plausibility based on its training up to that point. The closest corresponding timestamp we found is from December 2023 to January 2023.

\paragraph{Pythia 14M, 160M, and 2.8B}~\citep{biderman2023pythia} The Pythia models extend our exploration into the realms of modern architecture and model scalability, spanning the spectrum from an ultra-compact 14M to a robust 2.8B parameter setup. This inclusion allows us to assess plausibility inference capabilities across a broad range of model sizes, particularly emphasising the performance of smaller models like the Pythia 14M. The evaluation of Pythia models leverages their training data up to March 2020. The closest corresponding timestamp we found is from October 2020 to November 2020.

\paragraph{OLMo 1B and 7B}~\citep{groeneveld2024olmo} The newest models in our evaluation, OLMo's training data extends to March 2023. Their assessment against the \textsc{PRobELM} benchmark tests their ability to gauge plausibility based on a comprehensive and current dataset, without incorporating the very latest updates. The closest corresponding timestamp we found is from December 2023 to January 2023.

\section{Algorithm for Generating Less Plausible Scenarios}\label{app:less}
To systematically generate less plausible scenarios for the \textsc{PRobELM} benchmark, we use the algorithm shown in Algorithm \ref{alg:less}, which leverages the statistical distributions of entity co-occurrences within Wikidata. This method ensures that the generated scenarios, while being less plausible, still maintain logical coherence and relevance to the given context.

\input{algorithm/alg}

\section{Additional Results on Time-Specific Datasets} \label{sec:add_results}

We show the plausibility performance for models evaluated on \textsc{PRobELM} with different timestamps in \autoref{tab:201712_201801} (2017), \autoref{tab:202010_202011} (2020), \autoref{tab:202112_202201} (2021), and \autoref{tab:202311_202312} (2023).

\input{tables/2017}

\input{tables/2020}
\input{tables/2022}

\input{tables/11-12-2023}

\section{More details on \textsc{PRobELM}}\label{sec:datasets_detail}

An example of the prompts used to evaluate the models is shown in \autoref{tab:testingFormats}.

\input{tables/prompts}

We show the relations included in \textsc{PRobELM} and different types of prompts in \autoref{tab:relation_statement}, \autoref{tab:relation_tc}, and \autoref{tab:relation_qa}.

\input{tables/relations_statement}
\input{tables/relations_text_completion}
\input{tables/relations_qa}

\end{document}

%% file: tables/models.tex
\begin{table}[t]
\centering
\scalebox{0.85}{
\addtolength{\tabcolsep}{-2.5pt}
\renewcommand{\arraystretch}{1.1}
\begin{tabular}{@{}llcc@{}}
\toprule
\sc \textbf{Model (size)} & \textbf{Pre-training Data Up To} & \textbf{Evaluation Timestamps} & \textbf{Number of Scenarios} \\ 
\midrule
\textsc{GPT-2} (124M, 1.5B) & December 2017 & Dec 2017 - Jan 2018 & 9,328 \\
\textsc{BLOOM} (560M, 7B) & December 2021 & Dec 2021 - Jan 2022 & 8,569 \\ 
\textsc{LLaMA} (7B) & August 2022 & Nov 2023 - Dec 2023 & 5,280 \\
\textsc{Pythia} (14M, 160M, 2.8B) & March 2020 & Oct 2020 - Nov 2020 & 5,104 \\
\textsc{OLMo} (1B, 7B) & March 2023 & Nov 2023 - Dec 2023 & 5,280 \\
\bottomrule
\end{tabular}
}
\caption{Evaluated models and their corresponding data timelines in the \textsc{PRobELM} benchmark. This table highlights the diversity of model sizes, their training data recency, and the specific timestamps chosen for evaluating their plausibility inference capabilities.}
\label{tab:models}

\end{table}

%% file: tables/individual_timeframe.tex
\begin{table}[t]
\centering
\scalebox{0.85}{
\addtolength{\tabcolsep}{-1pt}
\renewcommand{\arraystretch}{1.2}

\begin{tabular}{lcccccccccc}
\toprule
 \multirow{2}{*}{\sc \textbf{ Model}}   & \multicolumn{1}{c}{\textbf{AVG}} 
& \multicolumn{3}{c}{ \textbf{Statement}} 
& \multicolumn{3}{c}{ \textbf{Text Completion}} 
& \multicolumn{3}{c}{\textbf{Question Answering}} 

\\
   & \sc plausibility 
   & \sc {acc.} & \sc  {mrr} & \sc  {ndcg}
   & \sc {acc.} & \sc  {mrr} & \sc  {ndcg}
   & \sc {acc.} & \sc  {mrr} & \sc  {ndcg}\\ 
\cmidrule(r){1-1} \cmidrule(r){2-2} \cmidrule(lr){3-5} \cmidrule(lr){6-8} \cmidrule(l){9-11} 
                         
\sc OLMo-1B &

45.07
& 29.96 & 49.92 & 50.13 
& 22.11 & 46.47 & 46.95
& 42.15 & 59.09 & 58.89

\\
\sc OLMo-7B & 

\underline{52.51}
& 31.20 & 51.62 & 51.57
& \textbf{44.62} & \underline{61.87} & \underline{61.43}
& \underline{45.45} & \underline{62.60} & \underline{62.23}

\\

\sc BLOOM-560M &

30.63
& 20.58 & 43.37 & 42.56
& ~~8.78 & 34.81 & 34.54
& 12.54 & 39.44 & 39.02

\\

\sc BLOOM-7B & 

40.74
& 27.35 & 49.90 & 49.50
& 17.44 & 44.21 & 43.39
& 12.80 & 41.11 & 40.29

\\

\sc GPT-2-124M & 

39.02
& 29.08 & 49.08 & 47.96
& 18.09 & 40.84 & 39.83
& 26.97 & 50.12 & 49.19

\\

\sc GPT-2-1.5B & 

45.51
& 29.19 & 49.81 & 48.83
& 25.52 & 48.92 & 48.11
& 41.95 & 58.92 & 58.34

\\
\sc Llama-7B & 

40.74
& 19.20 & 43.26 & 43.60 
& 29.96 & 52.08 & 51.81
 & 27.69 & 49.56 & 49.51 

 \\ 
\sc Pythia-14M & 

35.90
& 20.68 & 43.15 &  43.11
 & 17.67 & 39.48 & 39.37
 & 23.90 & 47.90 & 47.85 

 \\ 
\sc Pythia-160M & 

42.45
& \underline{34.54} & \underline{54.39} & \underline{55.03}
& 16.87 & 39.59 & 40.13
& 30.52 & 55.16 & 55.84 

\\
\sc Pythia-2.8B & 

\textbf{58.47}
& \textbf{43.78} & \textbf{61.17} & \textbf{61.14}
& \underline{43.57} & \textbf{62.77} & \textbf{62.33}
& \textbf{53.41} & \textbf{69.13} & \textbf{68.89} \\ 
\bottomrule

\end{tabular}
}
\caption{Evaluation results of language models on \textsc{PRobELM} with corresponding time-frame specific datasets. We report accuracy, MRR, and NDCG scores for three prompt types. AVG indicates average \textit{plausibility} across all prompts and metrics (average of nine scores). Best performances are in bold, and second-best results are underlined.}

\label{tab:main_results}
\end{table}

%% file: tables/compare_to_other_benchmarks.tex
\begin{table}[t]
\centering
\scalebox{0.8}{
\addtolength{\tabcolsep}{-1pt}
\renewcommand{\arraystretch}{1.2}

\begin{tabular}{lcccccc}
\toprule
 \multirow{2}{*}{\sc \textbf{ Model}}  

&   \textbf{\textsc{PRobELM}}
&  \textbf{\textsc{COPA}}
& \textbf{\textsc{ARC}}
& \textbf{\textsc{HellaSwag}}
& \textbf{\textsc{TruthfulQA}} 
& \textbf{\textsc{WinoGrande}} 
\\
& \sc {plausibility}  &  \sc{acc.}  &\sc{acc.-norm}  &\sc{acc.-norm} & \sc{acc.} & \sc{acc.}
\\

\midrule

\sc Pythia-2.8B 
 
& 58.47  \small{\textcolor{NavyBlue}{~~(1)}} 
& 79.0 \small{\textcolor{Maroon}{~~(4)}}  & 33.02   \small{\textcolor{Maroon}{~~(4)}} 
&59.30 \small{\textcolor{Maroon}{~~(5)}} 
& 35.88 \small{\textcolor{Maroon}{~~(7)}}
&59.12 \small{\textcolor{Maroon}{~~(5)}} 
 \\
 \sc OLMo-7B 

& 52.51 \small{\textcolor{NavyBlue}{~~(2)}} 
&  85.0 \small{\textcolor{DarkGreen}{~~(1)}} 
& 40.36  \small{\textcolor{NavyBlue}{~~(2)}}
&75.65  \small{\textcolor{NavyBlue}{~~(2)}}
&35.85  \small{\textcolor{Maroon}{~~(8)}}
& 66.38 \small{\textcolor{NavyBlue}{~~(2)}} 
\\
\sc GPT-2-1.5B

& 45.51 \small{\textcolor{NavyBlue}{~~(3)}} 
& 76.0 \small{\textcolor{Maroon}{~~(5)}} 
&  28.50 \small{\textcolor{Maroon}{~~(6)}} 
& 50.89 \small{\textcolor{Maroon}{~~(6)}} 
& 38.53 \small{\textcolor{Maroon}{~~(6)}}
&58.33 \small{\textcolor{Maroon}{~~(6)}} 
\\
\sc OLMo-1B 

& 45.07 \small{\textcolor{NavyBlue}{~~(4)}} 
  & 82.0 \small{\textcolor{DarkGreen}{~~(3)}} &31.06 \small{\textcolor{Maroon}{~~(5)}} 
  &62.92 \small{\textcolor{DarkGreen}{~~(3)}}
  &32.94 \small{\textcolor{Maroon}{~(10)}}
  & 59.98 \small{\textcolor{NavyBlue}{~~(4)}}

\\
\sc Pythia-160M 

& 42.45  \small{\textcolor{NavyBlue}{~~(5)}} 
 
& 64.0 \small{\textcolor{Maroon}{~~(7)}} 
& 23.21  \small{\textcolor{Maroon}{~~(8)}}
& 28.48 \small{\textcolor{Maroon}{~~(9)}}
& 44.43 \small{\textcolor{DarkGreen}{~~(2)}} 
& 49.88 \small{\textcolor{Maroon}{~(10)}}
\\
\sc LLaMA-7B 

& 40.74  \small{\textcolor{NavyBlue}{~~(6)}} 
 
&  85.0 \small{\textcolor{DarkGreen}{~~(1)}} & 44.62 \small{\textcolor{DarkGreen}{~~(1)}} 
&76.21  \small{\textcolor{DarkGreen}{~~(2)}}
&34.08 \small{\textcolor{Maroon}{~~(9)}}
& 70.01 \small{\textcolor{DarkGreen}{~~(1)}}
 \\
 \sc BLOOM-7B 

& 40.74  \small{\textcolor{NavyBlue}{~~(7)}} 

& 73.0 \small{\textcolor{DarkGreen}{~~(6)}} 
  
&33.45  \small{\textcolor{DarkGreen}{~~(3)}} 
&62.28 \small{\textcolor{DarkGreen}{~~(4)}}
&38.89 \small{\textcolor{DarkGreen}{~~(5)}}
&64.64 \small{\textcolor{DarkGreen}{~~(3)}}
\\
 \sc GPT-2-124M & 39.02 \small{\textcolor{NavyBlue}{~~(8)}} & 62.0 \small{\textcolor{Maroon}{~~(8)}} 
&22.70  \small{\textcolor{Maroon}{~~(9)}}
&31.14 \small{\textcolor{NavyBlue}{~~(8)}} 
 &   40.69 \small{\textcolor{DarkGreen}{~~(4)}}
 & 51.62 \small{\textcolor{DarkGreen}{~~(7)}} 
\\

\sc Pythia-14M 
  
&35.90  \small{\textcolor{NavyBlue}{~~(9)}} 

& 53.0  \small{\textcolor{Maroon}{~(10)}}
&21.25  \small{\textcolor{Maroon}{~(10)}}
& 26.12 \small{\textcolor{Maroon}{~(10)}}
&50.37   \small{\textcolor{DarkGreen}{~~(1)}} 
& 50.43 \small{\textcolor{NavyBlue}{~~(9)}} 

 \\ 
\sc BLOOM-560M 
& 30.63  \small{\textcolor{NavyBlue}{~(10)}} 
& 61.0 \small{\textcolor{DarkGreen}{~~(9)}} 
&23.81 \small{\textcolor{DarkGreen}{~~(7)}} 
&36.92 \small{\textcolor{DarkGreen}{~~(7)}} 
&42.43 \small{\textcolor{DarkGreen}{~~(3)}} 
& 51.38 \small{\textcolor{DarkGreen}{~~(8)}}
\\
\bottomrule

\end{tabular}
}
\caption{Language Models Performance on \textsc{PRobELM} versus other reasoning datasets. The \textbf{ranking} of the models is shown in brackets. ARC shows results on the challenge set. Same as before, \textit{plausibility} shows the average score of all prompt types and metrics (ACC., MRR, and NDCG). Models are ordered based on the plausibility rank on \textsc{PRobELM}.}

\label{tab:other_benchmarks}
\end{table}

%% file: algorithm/alg.tex
\begin{algorithm}

\begin{algorithmic} 
\Require{A triple from Wikidata $(s, r, o)$}
\Ensure{Ranked list of less plausible triples $L$}

    \State $r$ remains constant for all operations
    \State $P \gets \{p \mid (s, \text{subclass\_of}, p) \in \text{Wikidata}\}$
    \State $S \gets$ empty list to hold siblings of $s$
    \ForAll{entities $e$ sharing parent $P$}
      \If{$e \neq s$}
        \State Append $e$ to $S$
      \EndIf
    \EndFor
    \State $O_{all} \gets$ empty list to hold objects for all siblings
    \ForAll{$sib \in S$}
      \State $O_{sib} \gets$ all objects for triples $(sib, r, *)$
      \State Append $O_{sib}$ to $O_{all}$
    \EndFor
    \State Initialize an empty map $Freq$
    \ForAll{$obj \in O_{all}$}
      \If{$obj$ in $Freq$}
        \State $Freq[obj] \gets Freq[obj] + 1$
      \Else
        \State $Freq[obj] \gets 1$
      \EndIf
    \EndFor
    \State $L \gets \text{sort}(Freq, \text{key} = Freq.\text{get}, \text{reverse})$
    \State Repeat process with $o$ fixed to find subjects \\similar to $s$ and generate another ranked list $L$
    \caption{Generating Less Plausible Scenarios}
    \label{alg:less}
    \end{algorithmic}
    
\end{algorithm}

%% file: tables/2017.tex
\begin{table}[ht]
\centering
\scalebox{0.85}{
\addtolength{\tabcolsep}{-1pt}
\renewcommand{\arraystretch}{1.2}

\begin{tabular}{lcccccccccc}
\toprule
 \multirow{2}{*}{\sc \textbf{ Model}}   & \multicolumn{1}{c}{\textbf{AVG}} 
& \multicolumn{3}{c}{ \textbf{Statement}} 
& \multicolumn{3}{c}{ \textbf{Text Completion}} 
& \multicolumn{3}{c}{\textbf{Question Answering}} 

\\
   & \sc plausibility 
   & \sc {acc.} & \sc  {mrr} & \sc  {ndcg}
   & \sc {acc.} & \sc  {mrr} & \sc  {ndcg}
   & \sc {acc.} & \sc  {mrr} & \sc  {ndcg}\\ 
\cmidrule(r){1-1} \cmidrule(r){2-2} \cmidrule(lr){3-5} \cmidrule(lr){6-8} \cmidrule(l){9-11} 

\sc OLMo-1B & 46.64 (49.91)
& 30.52 & 48.90 & 47.57
& 26.75 & 49.72 & 48.78 
& \underline{44.28} & \underline{62.05} & \underline{61.21}
\\
\sc OLMo-7B & \textbf{54.68} (57.72)
& \underline{33.41} & \underline{52.70} &  \underline{51.76}
&  \textbf{46.73} & \textbf{64.68} & \textbf{64.37}
& \textbf{47.84} & \textbf{65.86} & \textbf{64.76}
\\

\sc BLOOM-560M & 37.95 (37.94)
& 23.20 &  46.05 & 45.35
& 20.76 &  43.82 & 43.40
& 24.31 & 47.67 & 47.01
\\
\sc BLOOM-7B & 42.48 (43.02)
& 29.86 & 50.01 &  49.39
& 28.86 & 52.22 &  51.78
& 23.31 & 48.72  &  48.18
\\

\sc GPT-2-124M & 39.02 (42.83)
& 29.08 & 49.08 &  47.96
& 18.09 & 40.84 & 39.83
& 26.97 & 50.12 & 49.19
\\

\sc GPT-2-1.5B & 45.51 (50.16)
& 29.19 &  49.81 &  48.83
& 25.52 &  48.92& 48.11
& 41.95 & 58.92 & 58.34
\\
\sc LLaMA-7B & 49.96 (48.38)
&  30.52 & 51.70 &  51.04
& \underline{40.95} & \underline{61.01} & \underline{60.37}
& 37.63 & 58.57 & 49.96
 \\ 
\sc Pythia-14M & 34.03 (37.77)
& 20.98 & 41.50 &  40.30
& 16.09 & 37.52 & 36.31
& 22.97 &  45.94& 44.65
 \\ 
\sc Pythia-160M & 39.69 (42.73)
& 30.30 & 50.27 &  49.51
& 15.54 & 37.08 & 36.15
& 30.97 & 53.75 & 53.62
\\
\sc Pythia-2.8B & \underline{50.85} (54.73)

& \textbf{34.30} & \textbf{53.08} &  \textbf{51.98}
& 36.51 & 58.44 & 58.37
& 42.51 & 61.69 & 60.80
 \\
\bottomrule
\end{tabular}
}
\caption{Evaluation Results of Language Models on \textsc{PRobELM} Using the Same Evaluation Dataset of December 2017 to January 2018. Plausibility Scores in Parenthesis Are Obtained Without Evaluating the ``Occupation'' Relation.} 

\label{tab:201712_201801}
\end{table}

%% file: tables/2020.tex
\begin{table}[ht]
\centering
\scalebox{0.85}{
\addtolength{\tabcolsep}{-1pt}
\renewcommand{\arraystretch}{1.2}

\begin{tabular}{lcccccccccc}
\toprule
 \multirow{2}{*}{\sc \textbf{ Model}}   & \multicolumn{1}{c}{\textbf{AVG}} 
& \multicolumn{3}{c}{ \textbf{Statement}} 
& \multicolumn{3}{c}{ \textbf{Text Completion}} 
& \multicolumn{3}{c}{\textbf{Question Answering}} 

\\
   & \sc plausibility 
   & \sc {acc.} & \sc  {mrr} & \sc  {ndcg}
   & \sc {acc.} & \sc  {mrr} & \sc  {ndcg}
   & \sc {acc.} & \sc  {mrr} & \sc  {ndcg}\\ 
\cmidrule(r){1-1} \cmidrule(r){2-2} \cmidrule(lr){3-5} \cmidrule(lr){6-8} \cmidrule(l){9-11} 

\sc OLMo-1B & 53.59
& 35.74 & 54.94 & 55.41
& 32.93 & 56.40 & 57.20
& 52.61 & 68.19 & \textbf{69.11}
\\
\sc OLMo-7B & \underline{58.09}
& \underline{39.36} & 57.53 & 57.25
& \textbf{49.60} & \textbf{67.22} & \textbf{66.31}
& 50.40 & 67.43 & 67.07
\\

\sc BLOOM-560M & 40.93
& 31.12 & 53.08 & 53.17
& 17.87 & 44.27 & 44.40
& 23.89 & 50.21 & 50.23
\\
\sc BLOOM-7B & 47.03
& 39.16 & \underline{58.70} & \underline{59.34}
& 28.71 & 54.87 & 55.23
& 23.69 & 51.60 & 51.81
\\

\sc GPT-2-124M & 42.96
& 32.73 & 52.78 & 52.99
& 18.47 & 43.58 & 43.72
& 30.72 & 55.54 & 56.11
\\

\sc GPT-2-1.5B & 52.36
& 37.35 & 57.67 & 57.74
& 26.70 & 52.22 & 52.36
& \underline{52.81} & \underline{68.37} & 68.61
\\
\sc LLaMA-7B & 47.14
& 25.30 & 48.46 & 48.69
& 35.74 & 57.89 & 58.25
& 34.33 & 57.61 & 57.95
 \\ 
\sc Pythia-14M & 35.89
& 20.68 & 43.15 & 43.11
& 17.67 & 39.48 & 39.37
& 23.90 & 47.90 & 47.85
 \\ 
\sc Pythia-160M & 42.45

& 34.54 & 54.39 & 55.03
& 16.87 & 39.59 & 40.13
& 30.52 & 55.16 & 55.84
\\
\sc Pythia-2.8B & \textbf{58.55}

& \textbf{43.78} &\textbf{ 61.17} & \textbf{61.14}
& \underline{43.57} & \underline{62.77} & \underline{62.33}
& \textbf{53.41} & \textbf{69.13} & \underline{68.89}
 \\
\bottomrule

\end{tabular}
}
\caption{Evaluation Results of Language Models on \textsc{PRobELM} Using the Same Evaluation Dataset of October to November 2020.}

\label{tab:202010_202011}
\end{table}

%% file: tables/2022.tex
\begin{table}[t]
\centering
\scalebox{0.85}{
\addtolength{\tabcolsep}{-1pt}
\renewcommand{\arraystretch}{1.2}

\begin{tabular}{lcccccccccc}
\toprule
 \multirow{2}{*}{\sc \textbf{ Model}}   & \multicolumn{1}{c}{\textbf{AVG}} 
& \multicolumn{3}{c}{ \textbf{Statement}} 
& \multicolumn{3}{c}{ \textbf{Text Completion}} 
& \multicolumn{3}{c}{\textbf{Question Answering}} 

\\
   & \sc plausibility 
   & \sc {acc.} & \sc  {mrr} & \sc  {ndcg}
   & \sc {acc.} & \sc  {mrr} & \sc  {ndcg}
   & \sc {acc.} & \sc  {mrr} & \sc  {ndcg}\\ 
\cmidrule(r){1-1} \cmidrule(r){2-2} \cmidrule(lr){3-5} \cmidrule(lr){6-8} \cmidrule(l){9-11} 

\sc OLMo-1B & 50.22
& \underline{38.52} & 56.12 &  55.17
  & 28.73 & 51.99 & 51.24
 & 45.92 & 63.21 & 62.12
\\
\sc OLMo-7B & \textbf{54.66}
& 38.02 & \underline{56.87} & \underline{55.83}
& \textbf{44.29} & \textbf{62.85} & \textbf{61.73} 
& \textbf{45.92} & \textbf{63.81} & \textbf{62.63}
\\

\sc BLOOM-560M & 30.63
& 20.58 & 43.37 & 42.56
& ~~8.78 & 34.81 & 34.54
& 12.54 & 39.44 & 39.02
\\
\sc BLOOM-7B & 36.22
& 27.35  & 49.90  & 49.50
& 17.44  & 44.21  & 43.39 
& 12.80  & 41.11  & 40.29 
\\

\sc GPT-2-124M & 38.92
& 32.25 & 51.89 & 51.22
& 14.81 & 39.64 & 39.38
& 24.22 & 48.82 & 48.25
\\

\sc GPT-2-1.5B & 46.68
& 35.51 &  56.02& 54.81
& 19.70  & 45.68 & 45.06
& 43.66 &  60.86&  59.86
\\
\sc LLaMA-7B & 48.14
 & 28.48 & 51.28 & 50.22
& 35.88 &  57.62& 56.14
 & 38.14 & 58.60 & 56.86
 \\ 
\sc Pythia-14M & 35.16
&20.08  & 42.90 &  43.15
 & 15.55& 38.66 & 39.38
&22.08 & 47.06 & 47.69
 \\ 
\sc Pythia-160M & 36.66
& 31.37  & 53.11  & 52.37
& 11.92  & 35.31  & 34.82
& 24.21  & 48.87  & 48.71
\\
\sc Pythia-2.8B & \underline{52.93}

&\textbf{41.53}  &  \textbf{58.64}& \textbf{57.78}
& 37.64 & 57.46 &56.60
& 44.54 & 61.86 &60.93
 \\
\bottomrule

\end{tabular}
}
\caption{Evaluation Results of Language Models on \textsc{PRobELM} Using the Same Evaluation Dataset of December 2021 to January 2022.}

\label{tab:202112_202201}
\end{table}

%% file: tables/11-12-2023.tex
\begin{table}[ht]
\centering
\scalebox{0.85}{
\addtolength{\tabcolsep}{-1pt}
\renewcommand{\arraystretch}{1.2}

\begin{tabular}{lcccccccccc}
\toprule
 \multirow{2}{*}{\sc \textbf{ Model}}   & \multicolumn{1}{c}{\textbf{AVG}} 
& \multicolumn{3}{c}{ \textbf{Statement}} 
& \multicolumn{3}{c}{ \textbf{Text Completion}} 
& \multicolumn{3}{c}{\textbf{Question Answering}} 

\\
   & \sc plausibility 
   & \sc {acc.} & \sc  {mrr} & \sc  {ndcg}
   & \sc {acc.} & \sc  {mrr} & \sc  {ndcg}
   & \sc {acc.} & \sc  {mrr} & \sc  {ndcg}\\ 
\cmidrule(r){1-1} \cmidrule(r){2-2} \cmidrule(lr){3-5} \cmidrule(lr){6-8} \cmidrule(l){9-11} 

\sc OLMo-1B & 45.07
& 29.96 & 49.92 &  50.13
  & 22.11 & 46.47 & 46.95
 & 42.15 & 59.09& 58.89
\\
\sc OLMo-7B & \textbf{52.51}
& \underline{31.20} & \underline{51.62} & \underline{51.57}
& \textbf{44.62} & \textbf{61.87} & \textbf{61.43} 
& \underline{45.45} & \underline{62.60} & \underline{62.23}
\\

\sc BLOOM-560M & 32.07
& 22.93 & 46.51 & 46.49
& ~~7.85 & 36.55 & 37.04
& 10.33 & 39.90 & 40.03
\\
\sc BLOOM-7B & 35.37
& 30.79 & 51.56 & 51.34
& 13.42 & 42.81 & 42.44
& ~~8.47 & 38.91 & 38.58
\\

\sc GPT-2-124M & 35.06
& 23.97 & 46.22 & 46.72
& 10.95 & 36.35 & 36.79
& 20.45 & 46.87 & 47.18
\\

\sc GPT-2-1.5B & 44.84
& 29.75 & 51.70 & 50.92
& 20.25 & 46.21 & 46.05
& 41.12 & 59.18 & 58.59 
\\
\sc LLaMA-7B & 40.77
 & 19.42 & 43.26 & 43.60
& 29.96 & 52.08 & 51.81
 & 27.69 & 49.56 & 49.51
 \\ 
\sc Pythia-14M & 28.89
& 12.40 & 37.21 &  37.74
 & ~~9.30& 32.76 & 33.77
& 14.67& 41.02 & 41.14
 \\ 
\sc Pythia-160M & 36.34

& 26.86 & 49.87 & 50.18
& 10.95 & 34.04 & 34.99
& 22.52& 48.60 & 49.05 
\\
\sc Pythia-2.8B & \underline{52.46}

& \textbf{36.16} & \textbf{55.55} &  \textbf{55.34}
 & \underline{36.98} & \underline{58.42} & \underline{58.26}
 & \textbf{46.07} & \textbf{63.45} & \textbf{62.86}
 \\
\bottomrule

\end{tabular}
}
\caption{Evaluation Results of Language Models on \textsc{PRobELM} Using the Same Evaluation Dataset of November to December 2023.}

\label{tab:202311_202312}
\end{table}

%% file: tables/prompts.tex
\begin{table}[ht]
\centering
\scalebox{0.9}{
\renewcommand{\arraystretch}{1.2}
\begin{tabular}{l|l}
\toprule
\textbf{Prompt Type} & \textbf{Examples} \\
\midrule
\sc Statement & ``Marcelo is Spanish.'', ``Marcelo is Irish.'', ``Marcelo is Chinese.'' \\
\sc Text Completion  & ``Fill in the blank: Marcelo has citizenship of \_\_\_\_. Answer: Spanish.''\\
\sc Question Answering & ``Question: What citizenship(s) does Marcelo have? Answer: Spanish.'' \\
\bottomrule
\end{tabular}
}
\caption{Illustrative Examples of Prompt Types Used in \textsc{PRobELM}.}
\label{tab:testingFormats}
\end{table}

%% file: tables/relations_statement.tex
\clearpage

\begin{table}[!ht]
\centering
\scalebox{0.75}{
\addtolength{\tabcolsep}{0pt}
\begin{tabular} {p{1cm}|p{5cm}|p{11cm}}
\toprule
\multicolumn{1}{l|}{\textsc{Pid}} & {\textsc{Relation}}                                         & \textsc{Template for \textbf{Statement}}    \\
\midrule

P5096 & member of the crew of&\textless subject\textgreater is a member of the crew of \textless object\textgreater.\\

P122 & basic form of government&The basic form of government of \textless subject\textgreater is \textless object\textgreater.\\

P3448 & stepparent&\textless subject\textgreater is the stepparent of \textless object\textgreater. \\
P1479 & has contributing factor&\textless object\textgreater is a contributing factor to \textless subject\textgreater.\\

P61 & discoverer or inventor&\textless subject\textgreater is the discoverer or inventor of \textless object\textgreater.\\

P3320 & board member&\textless subject\textgreater serves as a board member of \textless object\textgreater. \\
P7779 & member of military unit&\textless subject\textgreater is a member of the military unit \textless object\textgreater.\\

P98 & editor&\textless subject\textgreater is the editor of \textless object\textgreater.\\

P1411 & nominated for&\textless subject\textgreater was nominated for \textless object\textgreater.\\

P371 & presenter&\textless subject\textgreater is the presenter of \textless object\textgreater.\\

P1365 & replaces&\textless subject\textgreater replaces \textless object\textgreater.\\
P488 & chairperson&\textless subject\textgreater serves as the chairperson of \textless object\textgreater.\\
P27 & country of citizenship&\textless subject\textgreater's country of citizenship is \textless object\textgreater.\\
P20 & place of death&\textless subject\textgreater's place of death is \textless object\textgreater.\\
P1344 & participant in&\textless subject\textgreater was a participant in \textless object\textgreater.\\
P1366 & replaced by&\textless subject\textgreater was replaced by \textless object\textgreater.\\
P1412 & languages spoken written or signed&\textless subject\textgreater speaks the following languages: \textless object\textgreater.\\
P276 & location&The location of \textless subject\textgreater is \textless object\textgreater.\\
P407 & language of work or name&The language of \textless subject\textgreater's work or name is \textless object\textgreater.\\
P39 & position held&\textless subject\textgreater holds the position of \textless object\textgreater.\\
P1532 & country for sport&\textless subject\textgreater represents \textless object\textgreater in sports competitions.\\
P451 & unmarried partner&\textless subject\textgreater is the unmarried partner of \textless object\textgreater.\\
P54 & member of sports team&\textless subject\textgreater is a member of the sports team \textless object\textgreater.\\
P800 & notable work&\textless subject\textgreater's notable work includes \textless object\textgreater.\\
P551 & residence&\textless subject\textgreater's residence is in \textless object\textgreater.\\
P131 & located in the administrative territorial entity&\textless subject\textgreater is located in the administrative territorial entity \textless object\textgreater.\\
P106 & occupation&\textless subject\textgreater's occupation is \textless object\textgreater.\\
P69 & educated at&\textless subject\textgreater was educated at \textless object\textgreater.\\
P509 & cause of death&\textless subject\textgreater's cause of death was \textless object\textgreater.\\
P102 & member of political party&\textless subject\textgreater is a member of the political party \textless object\textgreater.\\
P19 & place of birth&\textless subject\textgreater's place of birth is \textless object\textgreater.\\
P115 & home venue&\textless subject\textgreater's home venue is \textless object\textgreater.\\
P1001 & applies to jurisdiction&\textless object\textgreater applies to the jurisdiction of \textless subject\textgreater.\\
P840 & narrative location&\textless subject\textgreater is set in the narrative location of \textless object\textgreater.\\
P108 & employer&\textless subject\textgreater is employed by \textless object\textgreater.\\
P57 & director&\textless subject\textgreater is the director of \textless object\textgreater.\\
P2416 & sports discipline competed in&\textless subject\textgreater competes in the sports discipline of \textless object\textgreater.\\
P400 & platform&\textless subject\textgreater is available on the platform \textless object\textgreater.\\
P1433 & published in&\textless subject\textgreater was published in \textless object\textgreater.\\
P1056 & product or material produced&\textless subject\textgreater produces \textless object\textgreater as a product or material.\\
P9071 & character type&\textless subject\textgreater is characterized as a \textless object\textgreater type.\\
P4100 & parliamentary group&\textless subject\textgreater is a member of the parliamentary group \textless object\textgreater.\\
P937 & work location&\textless subject\textgreater's work location is \textless object\textgreater.\\
P1066 & student of&\textless subject\textgreater is a student of \textless object\textgreater.\\
P1535 & used by&\textless object\textgreater is used by \textless subject\textgreater.\\
P6 & head of government&\textless subject\textgreater is the head of government of \textless object\textgreater.\\
P2283 & use&\textless subject\textgreater is used for \textless object\textgreater.\\
P812 & academic major&\textless subject\textgreater's academic major is \textless object\textgreater.\\
P1416 & affiliation&\textless subject\textgreater is affiliated with \textless object\textgreater.\\
P2522 & victory&\textless subject\textgreater achieved a victory in \textless object\textgreater.\\
P607 & conflict&\textless subject\textgreater was involved in the conflict \textless object\textgreater.\\
P749 & parent organization&\textless subject\textgreater is a part of the parent organization \textless object\textgreater.\\
P2283 & uses&\textless subject\textgreater uses \textless object\textgreater.\\
P802 & student&\textless subject\textgreater is a student at \textless object\textgreater.\\
P119 & place of burial&\textless subject\textgreater's place of burial is \textless object\textgreater.\\
P2842 & place of marriage&\textless subject\textgreater was married at \textless object\textgreater.\\
P286 & head coach&\textless subject\textgreater is the head coach of \textless object\textgreater.\\
P2541 & operating area&\textless subject\textgreater's operating area is \textless object\textgreater.\\
P1441 & present in work&\textless subject\textgreater is present in the work \textless object\textgreater.\\
P2650 & interested in&\textless subject\textgreater is interested in \textless object\textgreater.\\
P1027 & conferred by&\textless object\textgreater is conferred by \textless subject\textgreater.\\
P3300 & musical conductor&\textless subject\textgreater is the musical conductor of \textless object\textgreater.\\
P2715 & elected in&\textless subject\textgreater was elected in \textless object\textgreater.\\
P2937 & parliamentary term&\textless subject\textgreater served during the parliamentary term \textless object\textgreater.\\
P1399 & convicted of&\textless subject\textgreater was convicted of \textless object\textgreater.\\
P1686 & for work&\textless subject\textgreater is used for the work \textless object\textgreater.\\
P1196 & manner of death&\textless subject\textgreater's manner of death was \textless object\textgreater.\\
P2632 & place of detention&\textless subject\textgreater was detained at \textless object\textgreater.\\
P991 & successful candidate&\textless subject\textgreater was the successful candidate in \textless object\textgreater.\\
P2443 & stage reached&\textless subject\textgreater reached the stage \textless object\textgreater.\\
P6872 & has written for&\textless subject\textgreater has written for \textless object\textgreater.\\

\bottomrule
\end{tabular}
}
\caption{Relations and templates for \textit{statement} prompts used in \textsc{PRobELM}.}
\label{tab:relation_statement}
\end{table}

%% file: tables/relations_text_completion.tex
\clearpage

\begin{table}[!ht]
\centering
\scalebox{0.7}{
\addtolength{\tabcolsep}{0pt}
\begin{tabular} {p{1cm}|p{4cm}|p{13cm}}
\toprule
\multicolumn{1}{l|}{\textsc{Pid}} & {\textsc{Relation}}                                         & \textsc{Template for \textbf{Text Completion}}    \\
\midrule

P5096 & member of the crew of&Fill in the blank: \textless subject\textgreater is a member of the crew of \_\_\_\_. Answer:  \textless object\textgreater
\\

P122 & basic form of government&Fill in the blank: The basic form of government of \textless subject\textgreater is \_\_\_\_. Answer:  \textless object\textgreater
\\

P3448 & stepparent&Fill in the blank: \textless subject\textgreater is the stepparent of \_\_\_\_. Answer:  \textless object\textgreater
\\
P1479 & has contributing factor&Fill in the blank: \textless object\textgreater is a contributing factor to \_\_\_\_. Answer:  \textless subject\textgreater
\\

P61 & discoverer or inventor&Fill in the blank: \_\_\_\_ is the discoverer or inventor of \textless object\textgreater. Answer:  \textless subject\textgreater
\\

P3320 & board member&Fill in the blank: \textless subject\textgreater serves as a board member of \_\_\_\_. Answer:  \textless object\textgreater
  \\
P7779 & member of military unit&Fill in the blank: \textless subject\textgreater is a member of the military unit \_\_\_\_. Answer:  \textless object\textgreater
\\

P98 & editor&Fill in the blank: \_\_\_\_ is the editor of \textless object\textgreater. Answer:  \textless subject\textgreater
\\

P1411 & nominated for&Fill in the blank: \_\_\_\_ was nominated for \textless object\textgreater. Answer:  \textless subject\textgreater
\\

P371 & presenter&Fill in the blank: \_\_\_\_ is the presenter of \textless object\textgreater. Answer:  \textless subject\textgreater
\\

P1365 & replaces&Fill in the blank: \textless subject\textgreater replaces \_\_\_\_. Answer:  \textless object\textgreater
\\
P488 & chairperson&Fill in the blank: \_\_\_\_ serves as the chairperson of \textless object\textgreater. Answer:  \textless subject\textgreater
\\
P27 & country of citizenship&Fill in the blank: \textless subject\textgreater's country of citizenship is \_\_\_\_. Answer:  \textless object\textgreater
\\
P20 & place of death&Fill in the blank: \textless subject\textgreater's place of death is \_\_\_\_. Answer:  \textless object\textgreater
\\
P1344 & participant in&Fill in the blank: \_\_\_\_ was a participant in \textless object\textgreater. Answer:  \textless subject\textgreater
\\
P1366 & replaced by&Fill in the blank: \textless subject\textgreater was replaced by \_\_\_\_. Answer:  \textless object\textgreater
\\
P1412 & languages spoken written or signed&Fill in the blank: \textless subject\textgreater speaks the following languages: \_\_\_\_. Answer:  \textless object\textgreater
\\
P276 & location&Fill in the blank: The location of \textless subject\textgreater is \_\_\_\_. Answer:  \textless object\textgreater
\\
P407 & language of work or name&Fill in the blank: The language of \textless subject\textgreater's work or name is \_\_\_\_. Answer:  \textless object\textgreater
\\
P39 & position held&Fill in the blank: \textless subject\textgreater holds the position of \_\_\_\_. Answer:  \textless object\textgreater
\\
P1532 & country for sport&Fill in the blank: \_\_\_\_ represents \textless object\textgreater in sports competitions. Answer:  \textless subject\textgreater
\\
P451 & unmarried partner&Fill in the blank: \textless subject\textgreater is the unmarried partner of \_\_\_\_. Answer:  \textless object\textgreater
\\
P54 & member of sports team&Fill in the blank: \textless subject\textgreater is a member of the sports team \_\_\_\_. Answer:  \textless object\textgreater\\
P800 & notable work&Fill in the blank: \textless subject\textgreater's notable work includes \_\_\_\_. Answer:  \textless object\textgreater
\\
P551 & residence&Fill in the blank: \textless subject\textgreater's residence is in \_\_\_\_. Answer:  \textless object\textgreater
\\
P131 & located in the administrative territorial entity&Fill in the blank: \textless subject\textgreater is located in the administrative territorial entity \_\_\_\_. Answer:  \textless object\textgreater
\\
P106 & occupation&Fill in the blank: \textless subject\textgreater's occupation is \_\_\_\_. Answer:  \textless object\textgreater
\\
P69 & educated at&Fill in the blank: \textless subject\textgreater was educated at \_\_\_\_. Answer:  \textless object\textgreater
\\
P509 & cause of death&Fill in the blank: \textless subject\textgreater's cause of death was \_\_\_\_. Answer:  \textless object\textgreater
\\
P102 & member of political party&Fill in the blank: \textless subject\textgreater is a member of the political party \_\_\_\_. Answer:  \textless object\textgreater
\\
P19 & place of birth&Fill in the blank: \textless subject\textgreater's place of birth is \_\_\_\_. Answer:  \textless object\textgreater
\\
P115 & home venue&Fill in the blank: \textless subject\textgreater's home venue is \_\_\_\_. Answer:  \textless object\textgreater
\\
P1001 & applies to jurisdiction&Fill in the blank: \textless object\textgreater applies to the jurisdiction of \_\_\_\_. Answer:  \textless subject\textgreater
\\
P840 & narrative location&Fill in the blank: \textless subject\textgreater is set in the narrative location of \_\_\_\_. Answer:  \textless object\textgreater
\\
P108 & employer&Fill in the blank: \textless subject\textgreater is employed by \_\_\_\_. Answer:  \textless object\textgreater
\\
P57 & director&Fill in the blank: \_\_\_\_ is the director of \textless object\textgreater. Answer:  \textless subject\textgreater
\\
P2416 & sports discipline competed in&Fill in the blank: \textless subject\textgreater competes in the sports discipline of \_\_\_\_. Answer:  \textless object\textgreater
\\
P400 & platform&Fill in the blank: \textless subject\textgreater is available on the platform \_\_\_\_. Answer:  \textless object\textgreater
\\
P1433 & published in&Fill in the blank: \textless subject\textgreater was published in \_\_\_\_. Answer:  \textless object\textgreater
\\
P1056 & product or material produced&Fill in the blank: \textless subject\textgreater produces \_\_\_\_ as a product or material. Answer:  \textless object\textgreater
\\
P9071 & character type&Fill in the blank: \textless subject\textgreater is characterized as a \_\_\_\_ type. Answer:  \textless object\textgreater
\\
P4100 & parliamentary group&Fill in the blank: \textless subject\textgreater is a member of the parliamentary group \_\_\_\_. Answer:  \textless object\textgreater
\\
P937 & work location&Fill in the blank: \textless subject\textgreater's work location is \_\_\_\_. Answer:  \textless object\textgreater
\\
P1066 & student of&Fill in the blank: \textless subject\textgreater is a student of \_\_\_\_. Answer:  \textless object\textgreater
\\
P1535 & used by&Fill in the blank: \textless object\textgreater is used by \_\_\_\_. Answer:  \textless subject\textgreater
\\
P6 & head of government&Fill in the blank: \_\_\_\_ is the head of government of \textless object\textgreater. Answer:  \textless subject\textgreater\\
P2283 & use&Fill in the blank: \textless subject\textgreater is used for \_\_\_\_. Answer:  \textless object\textgreater
 \\
P812 & academic major&Fill in the blank: \textless subject\textgreater's academic major is \_\_\_\_. Answer:  \textless object\textgreater
\\
P1416 & affiliation&Fill in the blank: \textless subject\textgreater is affiliated with \_\_\_\_. Answer:  \textless object\textgreater
\\
P2522 & victory&Fill in the blank: \textless subject\textgreater achieved a victory in \_\_\_\_. Answer:  \textless object\textgreater
\\
P607 & conflict&Fill in the blank: \textless subject\textgreater was involved in the conflict \_\_\_\_. Answer:  \textless object\textgreater
\\
P749 & parent organization&Fill in the blank: \textless subject\textgreater is a part of the parent organization \_\_\_\_. Answer:  \textless object\textgreater
\\
P2283 & uses&Fill in the blank: \textless subject\textgreater uses \_\_\_\_. Answer:  \textless object\textgreater
\\
P802 & student&Fill in the blank: \textless subject\textgreater is a student at \_\_\_\_. Answer:  \textless object\textgreater
\\
P119 & place of burial&Fill in the blank: \textless subject\textgreater's place of burial is \_\_\_\_. Answer:  \textless object\textgreater
\\
P2842 & place of marriage&Fill in the blank: \textless subject\textgreater was married at \_\_\_\_. Answer:  \textless object\textgreater
\\
P286 & head coach&Fill in the blank: \_\_\_\_ is the head coach of \textless object\textgreater. Answer:  \textless subject\textgreater
\\
P2541 & operating area&Fill in the blank: \textless subject\textgreater's operating area is \_\_\_\_. Answer:  \textless object\textgreater
\\
P1441 & present in work&Fill in the blank: \textless subject\textgreater is present in the work \_\_\_\_. Answer:  \textless object\textgreater
\\
P2650 & interested in&Fill in the blank: \textless subject\textgreater is interested in \_\_\_\_. Answer:  \textless object\textgreater
\\
P1027 & conferred by&Fill in the blank: \textless object\textgreater is conferred by \_\_\_\_. Answer:  \textless subject\textgreater
\\
P3300 & musical conductor&Fill in the blank: \_\_\_\_ is the musical conductor of \textless object\textgreater. Answer:  \textless subject\textgreater
\\
P2715 & elected in&Fill in the blank: \textless subject\textgreater was elected in \textless object\textgreater. Answer:  \textless object\textgreater
\\
P2937 & parliamentary term&Fill in the blank: \textless subject\textgreater served during the parliamentary term \_\_\_\_. Answer:  \textless object\textgreater
\\
P1399 & convicted of&Fill in the blank: \textless subject\textgreater was convicted of \_\_\_\_. Answer:  \textless object\textgreater
\\
P1686 & for work&Fill in the blank: \textless subject\textgreater is used for the work \_\_\_\_. Answer:  \textless object\textgreater
\\
P1196 & manner of death&Fill in the blank: \textless subject\textgreater's manner of death was \_\_\_\_. Answer:  \textless object\textgreater
\\
P2632 & place of detention&Fill in the blank: \textless subject\textgreater was detained at \_\_\_\_. Answer:  \textless object\textgreater
\\
P991 & successful candidate&Fill in the blank: \textless subject\textgreater was the successful candidate in \_\_\_\_. Answer:  \textless object\textgreater
\\
P2443 & stage reached&Fill in the blank: \textless subject\textgreater reached the stage \_\_\_\_. Answer:  \textless object\textgreater
\\
P6872 & has written for&Fill in the blank: \textless subject\textgreater has written for \_\_\_\_. Answer:  \textless object\textgreater
\\

\bottomrule
\end{tabular}
}
\caption{Relations and templates for \textit{text completion} prompts used in \textsc{PRobELM}.}
\label{tab:relation_tc}
\end{table}

%% file: tables/relations_qa.tex
\clearpage

\begin{table}[!ht]
\centering
\scalebox{0.7}{
\addtolength{\tabcolsep}{0pt}
\begin{tabular} {p{1cm}|p{4cm}|p{13cm}}
\toprule
\multicolumn{1}{l|}{\textsc{Pid}} & {\textsc{Relation}}                                         & \textsc{Template for \textbf{Question Answering}}    \\
\midrule

P5096 & member of the crew of&  Question: \textless subject\textgreater is a member of the crew of what? Answer: \textless object\textgreater.
\\

P122 & basic form of government& Question: What is the basic form of government of \textless subject\textgreater? Answer: \textless object\textgreater.
\\

P3448 & stepparent&  Question: \textless subject\textgreater is the stepparent of who? Answer: \textless object\textgreater.
\\
P1479 & has contributing factor& Question: \textless object\textgreater is a contributing factor to What? Answer: \textless subject\textgreater.
\\

P61 & discoverer or inventor&  Question: Who is the discoverer or inventor of \textless object\textgreater? Answer: \textless subject\textgreater.
\\

P3320 & board member& Question: \textless subject\textgreater serves as a board member of what? Answer: \textless object\textgreater.
 \\
P7779 & member of military unit&  Question: \textless subject\textgreater is a member of the military unit of what? Answer: \textless object\textgreater.
\\

P98 & editor&  Question: Who is is the editor of \textless object\textgreater? Answer: \textless subject\textgreater.
\\

P1411 & nominated for&  Question: Who was nominated for \textless object\textgreater? Answer: \textless subject\textgreater.
\\

P371 & presenter&  Question: Who is the presenter of \textless object\textgreater? Answer: \textless subject\textgreater.
\\

P1365 & replaces&  Question: Whom did \textless subject\textgreater replace? Answer: \textless object\textgreater.
\\
P488 & chairperson&  Question: Who serves as the chairperson of \textless object\textgreater? Answer: \textless subject\textgreater.
\\
P27 & country of citizenship&  Question: What is \textless subject\textgreater's country of citizenship? Answer: \textless object\textgreater.
\\
P20 & place of death& Question: Where is \textless subject\textgreater's place of death? Answer: \textless object\textgreater.
\\
P1344 & participant in& Question: Who was a participant in \textless object\textgreater? Answer: \textless subject\textgreater.
\\
P1366 & replaced by& Question: What was \textless subject\textgreater replaced by? Answer: \textless object\textgreater.
\\
P1412 & languages spoken written or signed& Question: What language does \textless subject\textgreater speaks? Answer: \textless object\textgreater.
\\
P276 & location& Question: Where is the location of \textless subject\textgreater? Answer: \textless object\textgreater.
\\
P407 & language of work or name& Question: What is the language of \textless subject\textgreater's work or name? Answer: \textless object\textgreater.
\\
P39 & position held& Question: What position does \textless subject\textgreater hold? Answer: \textless object\textgreater.
\\
P1532 & country for sport& Question: Who represents \textless object\textgreater in sports competitions? Answer: \textless subject\textgreater.
\\
P451 & unmarried partner& Question: Who is the unmarried partner of \textless subject\textgreater? Answer: \textless object\textgreater.
\\
P54 & member of sports team& Question: \textless subject\textgreater is a member of which sports team? Answer: \textless object\textgreater.
\\
P800 & notable work& Question: What is a notable work of \textless subject\textgreater? Answer: \textless object\textgreater.
\\
P551 & residence& Question: Where is the residence of \textless subject\textgreater? Answer: \textless object\textgreater.
\\
P131 & located in the administrative territorial entity& Question: In which administrative territorial entity is \textless subject\textgreater located? Answer: \textless object\textgreater.
\\
P106 & occupation& Question: What is the occupation of \textless subject\textgreater? Answer: \textless object\textgreater.
\\
P69 & educated at& Question: Where was \textless subject\textgreater educated? Answer: \textless object\textgreater.
\\
P509 & cause of death& Question: What was the cause of death of \textless subject\textgreater? Answer: \textless object\textgreater.
\\
P102 & member of political party& Question: Which political party is \textless subject\textgreater a member of? Answer: \textless object\textgreater.
\\
P19 & place of birth& Question: Where is the place of birth of \textless subject\textgreater? Answer: \textless object\textgreater.
\\
P115 & home venue& Question: What is the home venue of \textless subject\textgreater? Answer: \textless object\textgreater.
\\
P1001 & applies to jurisdiction& Question: To which jurisdiction does \textless object\textgreater apply? Answer: \textless subject\textgreater.
\\
P840 & narrative location& Question: What is the narrative location of \textless subject\textgreater? Answer: \textless object\textgreater.
\\
P108 & employer& Question: Who is the employer of \textless subject\textgreater? Answer: \textless object\textgreater.
\\
P57 & director& Question: Who is the director of \textless object\textgreater? Answer: \textless subject\textgreater.
\\
P2416 & sports discipline competed in& Question: In which sports discipline does \textless subject\textgreater compete? Answer: \textless object\textgreater.
\\
P400 & platform& Question: On which platform is \textless subject\textgreater available? Answer: \textless object\textgreater.
\\
P1433 & published in& Question: Where was \textless subject\textgreater published? Answer: \textless object\textgreater.
\\
P1056 & product or material produced& Question: What product or material does \textless subject\textgreater produce? Answer: \textless object\textgreater.
\\
P9071 & character type& Question: What type of character is \textless subject\textgreater characterized as? Answer: \textless object\textgreater.
\\
P4100 & parliamentary group& Question: Which parliamentary group is \textless subject\textgreater a member of? Answer: \textless object\textgreater.
\\
P937 & work location& Question: What is the work location of \textless subject\textgreater? Answer: \textless object\textgreater.
\\
P1066 & student of& Question: Who is \textless subject\textgreater a student of? Answer: \textless object\textgreater.
\\
P1535 & used by& Question: Who uses \textless object\textgreater? Answer: \textless subject\textgreater.
\\
P6 & head of government& Question: Who is the head of government of \textless object\textgreater? Answer: \textless subject\textgreater.
\\
P2283 & use& Question: What is \textless subject\textgreater used for? Answer: \textless object\textgreater.
\\
P812 & academic major& Question: What is \textless subject\textgreater's academic major?Answer: \textless object\textgreater.
\\
P1416 & affiliation& Question: What is \textless subject\textgreater affiliated with? Answer: \textless object\textgreater.
\\
P2522 & victory& Question: In which \textless object\textgreater did \textless subject\textgreater achieve a victory? Answer: \textless object\textgreater.
\\
P607 & conflict& Question: In which conflict was \textless subject\textgreater involved? Answer: \textless object\textgreater.
\\
P749 & parent organization& Question: What parent organization is \textless subject\textgreater a part of? Answer: \textless object\textgreater.
\\
P2283 & uses& Question: What does \textless subject\textgreater use? Answer: \textless object\textgreater.
\\
P802 & student& Question: At which institution is \textless subject\textgreater a student? Answer: \textless object\textgreater.
\\
P119 & place of burial& Question: Where is \textless subject\textgreater buried? Answer: \textless object\textgreater.
\\
P2842 & place of marriage& Question: Where was \textless subject\textgreater married? Answer: \textless object\textgreater.
\\
P286 & head coach& Question: Who is the head coach of \textless object\textgreater? Answer: \textless subject\textgreater.
\\
P2541 & operating area& Question: What is the operating area of \textless subject\textgreater? Answer: \textless object\textgreater.
\\
P1441 & present in work& Question: In which work is \textless subject\textgreater present? Answer: \textless object\textgreater.
\\
P2650 & interested in&  Question: What is \textless subject\textgreater interested in? Answer: \textless object\textgreater.

\\
P1027 & conferred by& Question: Who conferred \textless object\textgreater? Answer: \textless subject\textgreater.
\\
P3300 & musical conductor& Question: Who is the musical conductor of \textless object\textgreater? Answer: \textless subject\textgreater.
\\
P2715 & elected in& Question: In what was \textless subject\textgreater elected? Answer: \textless object\textgreater.
\\
P2937 & parliamentary term& Question: During which parliamentary term did \textless subject\textgreater serve? Answer: \textless object\textgreater.
\\
P1399 & convicted of& Question: What was \textless subject\textgreater convicted of? Answer: \textless object\textgreater.
\\
P1686 & for work& Question: What is \textless subject\textgreater used for in terms of work? Answer: \textless object\textgreater.
\\
P1196 & manner of death& Question: What was the manner of death of \textless subject\textgreater? Answer: \textless object\textgreater.
\\
P2632 & place of detention& Question: Where was \textless subject\textgreater detained? Answer: \textless object\textgreater.
\\
P991 & successful candidate& Question: In which election or selection was \textless subject\textgreater the successful candidate? Answer: \textless object\textgreater.
\\
P2443 & stage reached& Question: What stage did \textless subject\textgreater reach? Answer: \textless object\textgreater.
\\
P6872 & has written for& Question: For whom or what has \textless subject\textgreater written? Answer: \textless object\textgreater.
\\

\bottomrule
\end{tabular}
}
\caption{Relations and templates for \textit{question answering} prompts used in \textsc{PRobELM}.}
\label{tab:relation_qa}
\end{table}

%% file: main.bbl
\begin{thebibliography}{28}
\providecommand{\natexlab}[1]{#1}
\providecommand{\url}[1]{\texttt{#1}}
\expandafter\ifx\csname urlstyle\endcsname\relax
  \providecommand{\doi}[1]{doi: #1}\else
  \providecommand{\doi}{doi: \begingroup \urlstyle{rm}\Url}\fi

\bibitem[Anil et~al.(2023)Anil, Dai, Firat, Johnson, Lepikhin, Passos, Shakeri, Taropa, Bailey, Chen, et~al.]{anil2023palm}
Rohan Anil, Andrew~M Dai, Orhan Firat, Melvin Johnson, Dmitry Lepikhin, Alexandre Passos, Siamak Shakeri, Emanuel Taropa, Paige Bailey, Zhifeng Chen, et~al.
\newblock {PaLM 2 Technical Report}.
\newblock \emph{arXiv preprint arXiv:2305.10403}, 2023.
\newblock URL \url{https://arxiv.org/abs/2305.10403}.

\bibitem[Biderman et~al.(2023)Biderman, Schoelkopf, Anthony, Bradley, O’Brien, Hallahan, Khan, Purohit, Prashanth, Raff, et~al.]{biderman2023pythia}
Stella Biderman, Hailey Schoelkopf, Quentin~Gregory Anthony, Herbie Bradley, Kyle O’Brien, Eric Hallahan, Mohammad~Aflah Khan, Shivanshu Purohit, USVSN~Sai Prashanth, Edward Raff, et~al.
\newblock {Pythia: A Suite for Analyzing Large Language Models Across Training and Scaling}.
\newblock In \emph{International Conference on Machine Learning}, pp.\  2397--2430. PMLR, 2023.
\newblock URL \url{https://proceedings.mlr.press/v202/biderman23a/biderman23a.pdf}.

\bibitem[Chowdhery et~al.(2023)Chowdhery, Narang, Devlin, Bosma, Mishra, Roberts, Barham, Chung, Sutton, Gehrmann, et~al.]{chowdhery2023palm}
Aakanksha Chowdhery, Sharan Narang, Jacob Devlin, Maarten Bosma, Gaurav Mishra, Adam Roberts, Paul Barham, Hyung~Won Chung, Charles Sutton, Sebastian Gehrmann, et~al.
\newblock {PaLM: Scaling Language Modeling with Pathways}.
\newblock \emph{Journal of Machine Learning Research}, 24\penalty0 (240):\penalty0 1--113, 2023.
\newblock URL \url{https://jmlr.org/papers/volume24/22-1144/22-1144.pdf}.

\bibitem[Clark et~al.(2018)Clark, Cowhey, Etzioni, Khot, Sabharwal, Schoenick, and Tafjord]{clark2018think}
Peter Clark, Isaac Cowhey, Oren Etzioni, Tushar Khot, Ashish Sabharwal, Carissa Schoenick, and Oyvind Tafjord.
\newblock {Think You Have Solved Question Answering? Try ARC, the AI2 Reasoning Challenge}.
\newblock \emph{arXiv preprint arXiv:1803.05457}, 2018.
\newblock URL \url{https://arxiv.org/abs/1803.05457}.

\bibitem[Gopalakrishnan et~al.(2019)Gopalakrishnan, Jha, Jin, and Zhang]{GOPALAKRISHNAN2019103141}
Vishrawas Gopalakrishnan, Kishlay Jha, Wei Jin, and Aidong Zhang.
\newblock {A Survey on Literature Based Discovery Approaches in Biomedical Domain}.
\newblock \emph{Journal of Biomedical Informatics}, 93:\penalty0 103141, 2019.
\newblock ISSN 1532-0464.
\newblock \doi{https://doi.org/10.1016/j.jbi.2019.103141}.
\newblock URL \url{https://www.sciencedirect.com/science/article/pii/S1532046419300590}.

\bibitem[Groeneveld et~al.(2024)Groeneveld, Beltagy, Walsh, Bhagia, Kinney, Tafjord, Jha, Ivison, Magnusson, Wang, et~al.]{groeneveld2024olmo}
Dirk Groeneveld, Iz~Beltagy, Pete Walsh, Akshita Bhagia, Rodney Kinney, Oyvind Tafjord, Ananya~Harsh Jha, Hamish Ivison, Ian Magnusson, Yizhong Wang, et~al.
\newblock {OLMo: Accelerating the Science of Language Models}.
\newblock \emph{arXiv preprint arXiv:2402.00838}, 2024.
\newblock URL \url{https://arxiv.org/abs/2402.00838}.

\bibitem[Gruver et~al.(2023)Gruver, Finzi, Qiu, and Wilson]{gruver2023large}
Nate Gruver, Marc~Anton Finzi, Shikai Qiu, and Andrew~Gordon Wilson.
\newblock {Large Language Models Are Zero-Shot Time Series Forecasters}.
\newblock In \emph{Thirty-seventh Conference on Neural Information Processing Systems}, 2023.
\newblock URL \url{https://openreview.net/forum?id=md68e8iZK1}.

\bibitem[Hao et~al.(2023)Hao, Gu, Ma, Hong, Wang, Wang, and Hu]{hao-etal-2023-reasoning}
Shibo Hao, Yi~Gu, Haodi Ma, Joshua Hong, Zhen Wang, Daisy Wang, and Zhiting Hu.
\newblock {Reasoning with Language Model is Planning with World Model}.
\newblock In Houda Bouamor, Juan Pino, and Kalika Bali (eds.), \emph{Proceedings of the 2023 Conference on Empirical Methods in Natural Language Processing}, pp.\  8154--8173, Singapore, December 2023. Association for Computational Linguistics.
\newblock \doi{10.18653/v1/2023.emnlp-main.507}.
\newblock URL \url{https://aclanthology.org/2023.emnlp-main.507}.

\bibitem[Jang et~al.(2022)Jang, Ye, Lee, Yang, Shin, Han, Kim, and Seo]{jang-etal-2022-temporalwiki}
Joel Jang, Seonghyeon Ye, Changho Lee, Sohee Yang, Joongbo Shin, Janghoon Han, Gyeonghun Kim, and Minjoon Seo.
\newblock {{T}emporal{W}iki: A Lifelong Benchmark for Training and Evaluating Ever-Evolving Language Models}.
\newblock In Yoav Goldberg, Zornitsa Kozareva, and Yue Zhang (eds.), \emph{Proceedings of the 2022 Conference on Empirical Methods in Natural Language Processing}, pp.\  6237--6250, Abu Dhabi, United Arab Emirates, December 2022. Association for Computational Linguistics.
\newblock \doi{10.18653/v1/2022.emnlp-main.418}.
\newblock URL \url{https://aclanthology.org/2022.emnlp-main.418}.

\bibitem[Ji et~al.(2023)Ji, Lee, Frieske, Yu, Su, Xu, Ishii, Bang, Madotto, and Fung]{ji2023survey}
Ziwei Ji, Nayeon Lee, Rita Frieske, Tiezheng Yu, Dan Su, Yan Xu, Etsuko Ishii, Ye~Jin Bang, Andrea Madotto, and Pascale Fung.
\newblock {Survey of Hallucination in Natural Language Generation}.
\newblock \emph{ACM Computing Surveys}, 55\penalty0 (12):\penalty0 1--38, 2023.
\newblock URL \url{https://dl.acm.org/doi/10.1145/3571730}.

\bibitem[Kim et~al.(2023)Kim, Sclar, Zhou, Bras, Kim, Choi, and Sap]{kim-etal-2023-fantom}
Hyunwoo Kim, Melanie Sclar, Xuhui Zhou, Ronan Bras, Gunhee Kim, Yejin Choi, and Maarten Sap.
\newblock {{FANT}o{M}: A Benchmark for Stress-testing Machine Theory of Mind in Interactions}.
\newblock In Houda Bouamor, Juan Pino, and Kalika Bali (eds.), \emph{Proceedings of the 2023 Conference on Empirical Methods in Natural Language Processing}, pp.\  14397--14413, Singapore, December 2023. Association for Computational Linguistics.
\newblock \doi{10.18653/v1/2023.emnlp-main.890}.
\newblock URL \url{https://aclanthology.org/2023.emnlp-main.890}.

\bibitem[Li et~al.(2023)Li, Cheng, Zhao, Nie, and Wen]{li-etal-2023-halueval}
Junyi Li, Xiaoxue Cheng, Xin Zhao, Jian-Yun Nie, and Ji-Rong Wen.
\newblock {{H}alu{E}val: A Large-Scale Hallucination Evaluation Benchmark for Large Language Models}.
\newblock In Houda Bouamor, Juan Pino, and Kalika Bali (eds.), \emph{Proceedings of the 2023 Conference on Empirical Methods in Natural Language Processing}, pp.\  6449--6464, Singapore, December 2023. Association for Computational Linguistics.
\newblock \doi{10.18653/v1/2023.emnlp-main.397}.
\newblock URL \url{https://aclanthology.org/2023.emnlp-main.397}.

\bibitem[Lin et~al.(2022)Lin, Hilton, and Evans]{lin-etal-2022-truthfulqa}
Stephanie Lin, Jacob Hilton, and Owain Evans.
\newblock {{T}ruthful{QA}: Measuring How Models Mimic Human Falsehoods}.
\newblock In Smaranda Muresan, Preslav Nakov, and Aline Villavicencio (eds.), \emph{Proceedings of the 60th Annual Meeting of the Association for Computational Linguistics (Volume 1: Long Papers)}, pp.\  3214--3252, Dublin, Ireland, May 2022. Association for Computational Linguistics.
\newblock \doi{10.18653/v1/2022.acl-long.229}.
\newblock URL \url{https://aclanthology.org/2022.acl-long.229}.

\bibitem[Min et~al.(2023)Min, Krishna, Lyu, Lewis, Yih, Koh, Iyyer, Zettlemoyer, and Hajishirzi]{min-etal-2023-factscore}
Sewon Min, Kalpesh Krishna, Xinxi Lyu, Mike Lewis, Wen-tau Yih, Pang Koh, Mohit Iyyer, Luke Zettlemoyer, and Hannaneh Hajishirzi.
\newblock {{FA}ct{S}core: Fine-grained Atomic Evaluation of Factual Precision in Long Form Text Generation}.
\newblock In Houda Bouamor, Juan Pino, and Kalika Bali (eds.), \emph{Proceedings of the 2023 Conference on Empirical Methods in Natural Language Processing}, pp.\  12076--12100, Singapore, December 2023. Association for Computational Linguistics.
\newblock \doi{10.18653/v1/2023.emnlp-main.741}.
\newblock URL \url{https://aclanthology.org/2023.emnlp-main.741}.

\bibitem[M{\"u}ndler et~al.(2024)M{\"u}ndler, He, Jenko, and Vechev]{mundler2024selfcontradictory}
Niels M{\"u}ndler, Jingxuan He, Slobodan Jenko, and Martin Vechev.
\newblock {Self-contradictory Hallucinations of Large Language Models: Evaluation, Detection and Mitigation}.
\newblock In \emph{The Twelfth International Conference on Learning Representations}, 2024.
\newblock URL \url{https://openreview.net/forum?id=EmQSOi1X2f}.

\bibitem[Radford et~al.(2019)Radford, Wu, Child, Luan, Amodei, Sutskever, et~al.]{radford2019language}
Alec Radford, Jeffrey Wu, Rewon Child, David Luan, Dario Amodei, Ilya Sutskever, et~al.
\newblock {Language Models are Unsupervised Multitask Learners}.
\newblock \emph{OpenAI blog}, 1\penalty0 (8):\penalty0 9, 2019.

\bibitem[Roemmele et~al.(2011)Roemmele, Bejan, and Gordon]{roemmele2011choice}
Melissa Roemmele, Cosmin~Adrian Bejan, and Andrew~S Gordon.
\newblock {Choice of Plausible alternatives: An Evaluation of Commonsense Causal Reasoning}.
\newblock In \emph{2011 AAAI Spring Symposium Series}, 2011.
\newblock URL \url{https://cdn.aaai.org/ocs/2418/2418-10878-1-PB.pdf}.

\bibitem[Sakaguchi et~al.(2021)Sakaguchi, Bras, Bhagavatula, and Choi]{sakaguchi2021winogrande}
Keisuke Sakaguchi, Ronan~Le Bras, Chandra Bhagavatula, and Yejin Choi.
\newblock {Winogrande: An Adversarial Winograd Schema Challenge at Scale}.
\newblock \emph{Communications of the ACM}, 64\penalty0 (9):\penalty0 99--106, 2021.
\newblock URL \url{https://dl.acm.org/doi/10.1145/3474381}.

\bibitem[Scao et~al.(2023)Scao, Fan, Akiki, Pavlick, Ili{\'c}, Hesslow, Castagn{\'e}, Luccioni, Yvon, et~al.]{scao:hal-03850124}
Teven~Le Scao, Angela Fan, Christopher Akiki, Ellie Pavlick, Suzana Ili{\'c}, Daniel Hesslow, Roman Castagn{\'e}, Alexandra~Sasha Luccioni, Fran{\c c}ois Yvon, et~al.
\newblock {BLOOM: A 176B-Parameter Open-Access Multilingual Language Model}.
\newblock working paper or preprint, November 2023.
\newblock URL \url{https://inria.hal.science/hal-03850124}.

\bibitem[Touvron et~al.(2023)Touvron, Lavril, Izacard, Martinet, Lachaux, Lacroix, Rozi{\`e}re, Goyal, Hambro, Azhar, et~al.]{touvron2023llama}
Hugo Touvron, Thibaut Lavril, Gautier Izacard, Xavier Martinet, Marie-Anne Lachaux, Timoth{\'e}e Lacroix, Baptiste Rozi{\`e}re, Naman Goyal, Eric Hambro, Faisal Azhar, et~al.
\newblock {Llama: Open and Efficient Foundation Language Models}.
\newblock \emph{arXiv preprint arXiv:2302.13971}, 2023.
\newblock URL \url{https://arxiv.org/abs/2302.13971}.

\bibitem[Wang et~al.(2018)Wang, Singh, Michael, Hill, Levy, and Bowman]{wang-etal-2018-glue}
Alex Wang, Amanpreet Singh, Julian Michael, Felix Hill, Omer Levy, and Samuel Bowman.
\newblock {{GLUE}: A Multi-Task Benchmark and Analysis Platform for Natural Language Understanding}.
\newblock In Tal Linzen, Grzegorz Chrupa{\l}a, and Afra Alishahi (eds.), \emph{Proceedings of the 2018 {EMNLP} Workshop {B}lackbox{NLP}: Analyzing and Interpreting Neural Networks for {NLP}}, pp.\  353--355, Brussels, Belgium, November 2018. Association for Computational Linguistics.
\newblock \doi{10.18653/v1/W18-5446}.
\newblock URL \url{https://aclanthology.org/W18-5446}.

\bibitem[Wang et~al.(2019)Wang, Pruksachatkun, Nangia, Singh, Michael, Hill, Levy, and Bowman]{wang2019superglue}
Alex Wang, Yada Pruksachatkun, Nikita Nangia, Amanpreet Singh, Julian Michael, Felix Hill, Omer Levy, and Samuel Bowman.
\newblock Superglue: A stickier benchmark for general-purpose language understanding systems.
\newblock \emph{Advances in neural information processing systems}, 32, 2019.
\newblock URL \url{https://proceedings.neurips.cc/paper_files/paper/2019/file/4496bf24afe7fab6f046bf4923da8de6-Paper.pdf}.

\bibitem[Wang et~al.(2023)Wang, Downey, Ji, and Hope]{wang2023learning}
Qingyun Wang, Doug Downey, Heng Ji, and Tom Hope.
\newblock {SciMON: Scientific Inspiration Machines Optimized for Novelty}.
\newblock \emph{arXiv preprint arXiv:2305.14259}, 2023.
\newblock URL \url{https://arxiv.org/abs/2305.14259}.

\bibitem[Whitehouse et~al.(2023)Whitehouse, Vania, Aji, Christodoulopoulos, and Pierleoni]{webie}
Chenxi Whitehouse, Clara Vania, Alham~Fikri Aji, Christos Christodoulopoulos, and Andrea Pierleoni.
\newblock {{W}eb{IE}: Faithful and Robust Information Extraction on the Web}.
\newblock In Anna Rogers, Jordan Boyd-Graber, and Naoaki Okazaki (eds.), \emph{Proceedings of the 61st Annual Meeting of the Association for Computational Linguistics (Volume 1: Long Papers)}, pp.\  7734--7755, Toronto, Canada, July 2023. Association for Computational Linguistics.
\newblock \doi{10.18653/v1/2023.acl-long.428}.
\newblock URL \url{https://aclanthology.org/2023.acl-long.428}.

\bibitem[Xu et~al.(2024)Xu, Jain, and Kankanhalli]{xu2024hallucination}
Ziwei Xu, Sanjay Jain, and Mohan Kankanhalli.
\newblock {Hallucination is Inevitable: An Innate Limitation of Large Language Models}.
\newblock \emph{arXiv preprint arXiv:2401.11817}, 2024.
\newblock URL \url{https://arxiv.org/abs/2401.11817}.

\bibitem[Ye et~al.(2024)Ye, Yang, Pang, Wang, Wong, Yilmaz, Shi, and Tu]{ye2024benchmarking}
Fanghua Ye, Mingming Yang, Jianhui Pang, Longyue Wang, Derek~F Wong, Emine Yilmaz, Shuming Shi, and Zhaopeng Tu.
\newblock {Benchmarking LLMs via Uncertainty Quantification}.
\newblock \emph{arXiv preprint arXiv:2401.12794}, 2024.
\newblock URL \url{https://arxiv.org/abs/2401.12794}.

\bibitem[Yu et~al.(2023)Yu, Wang, Tu, Cao, Zhang-Li, Lv, Peng, Yao, Zhang, Li, et~al.]{yu2023kola}
Jifan Yu, Xiaozhi Wang, Shangqing Tu, Shulin Cao, Daniel Zhang-Li, Xin Lv, Hao Peng, Zijun Yao, Xiaohan Zhang, Hanming Li, et~al.
\newblock {KoLA: Carefully Benchmarking World Knowledge of Large Language Models}.
\newblock \emph{arXiv preprint arXiv:2306.09296}, 2023.
\newblock URL \url{https://arxiv.org/abs/2306.09296}.

\bibitem[Zellers et~al.(2019)Zellers, Holtzman, Bisk, Farhadi, and Choi]{zellers-etal-2019-hellaswag}
Rowan Zellers, Ari Holtzman, Yonatan Bisk, Ali Farhadi, and Yejin Choi.
\newblock {{H}ella{S}wag: Can a Machine Really Finish Your Sentence?}
\newblock In Anna Korhonen, David Traum, and Llu{\'\i}s M{\`a}rquez (eds.), \emph{Proceedings of the 57th Annual Meeting of the Association for Computational Linguistics}, pp.\  4791--4800, Florence, Italy, July 2019. Association for Computational Linguistics.
\newblock \doi{10.18653/v1/P19-1472}.
\newblock URL \url{https://aclanthology.org/P19-1472}.

\end{thebibliography}
